\documentclass[journal,twoside,web]{ieeecolor}
\usepackage{tmi}
\usepackage{cite}
\usepackage{amsmath,amssymb,amsfonts}
\usepackage{algorithmic}
\usepackage{graphicx}
\usepackage{textcomp}
\usepackage{booktabs}
\usepackage{newfloat}
\usepackage{listings}
\usepackage[most]{tcolorbox}
\usepackage[table,dvipsnames]{xcolor}

\definecolor{p3}{RGB}{40,120,181}
\usepackage{amssymb}
\usepackage{tikz}
\usepackage{pifont}
\usepackage{xcolor}
\usepackage[table]{xcolor}
\usepackage{bm}
\usepackage{subcaption}
\makeatletter
\let\NAT@parse\undefined
\makeatother
\usepackage{hyperref}
\usepackage{microtype}

\hypersetup{
    colorlinks=true,   
    linkcolor=blue,    
    filecolor=magenta, 
    urlcolor=cyan,    
    citecolor=blue,   
    pdftitle={Your Document Title},
    pdfauthor={Your Name},
    pdfsubject={Subject},
    pdfkeywords={keyword1, keyword2}
}
\def\BibTeX{{\rm B\kern-.05em{\sc i\kern-.025em b}\kern-.08em
    T\kern-.1667em\lower.7ex\hbox{E}\kern-.125emX}}
{Zhang \MakeLowercase{\textit{et al.}}: Optimize Surgical Triplet Recognition: A Knowledge-Driven Mixture-of-Experts Solution}

\definecolor{mediumred}{RGB}{200,0,0}

\DeclareRobustCommand{\redcircle}{%
  \tikz[baseline=-0.5ex]\filldraw[
    fill=gray!10,
    draw=mediumred,
    line width=0.7pt
  ] (0,0) circle (0.6ex);
}

\definecolor{mediumblue}{RGB}{0,90,200}
\DeclareRobustCommand{\bluecircle}{%
  \tikz[baseline=-0.5ex]\filldraw[
    fill=gray!10,
    draw=mediumblue,
    line width=0.7pt
  ] (0,0) circle (0.6ex);
}

\begin{document}
\title{Optimize Surgical Triplet Recognition: A Knowledge-Driven Mixture-of-Experts Solution}
\author{Yiyi Zhang, Yuchen Yuan, Ying Zheng, Jialun Pei, \IEEEmembership{Member, IEEE}, Jinpeng Li, \\
Zheng Li,  \IEEEmembership{Senior Member, IEEE},
 and Pheng-Ann Heng, \IEEEmembership{Senior Member, IEEE}
\thanks{{This work was supported in part by the Research Grants Council of the Hong Kong Special Administrative Region (T45-401/22-N, C4042-23GF, 14214322, 14200623, 14203424, 14206325), the Hong Kong Innovation and Technology Fund (GHP/252/23SZ), the National Natural Science Foundation of China (62106236); and supported by the AIR@InnoHK Multiscale Medical Robotics Center, the Chow Yuk Ho Technology Center for Innovative Medicine, and the Li Ka Shing Institute of Health Sciences.} (Corresponding author: Jinpeng Li.)}
\thanks{Y. Zhang, Y. Yuan, J. Pei, J. Li, and P.-A. Heng are with the Department of Computer Science and Engineering, The Chinese University of Hong Kong, Hong Kong, China. P.-A. Heng is also with the Institute of Medical Intelligence and XR, The Chinese University of Hong Kong. (e-mail: yyzhang24@cse.cuhk.edu.hk; ycyuan22@cse.cuhk.edu.hk; peijialun@gmail.com; jpli21@cse.cuhk.edu.hk; pheng@cse.cuhk.edu.hk).}
\thanks{Y. Zheng is with the Department of Electrical and Electronic Engineering, The Hong Kong Polytechnic University, Hong Kong, China. (e-mail: ying1.zheng@polyu.edu.hk).}
\thanks{Z. Li is with the Department of Surgery, The Chinese University of Hong Kong, Hong Kong, China. (e-mail: lizheng@cuhk.edu.hk).}
}

\maketitle

\begin{abstract}
Surgical action triplet recognition constitutes a critical task in context-aware robot-assisted surgery, facilitating automatic surgical action perception by identifying instrument, verb, target, and their association. However, existing works struggle to analyze such complex surgical scenes due to three main issues: (1) component-level optimization conflicts caused by entangled feature spaces, (2) category-level optimization conflicts arising from severe data imbalance, and (3) lack of domain knowledge guidance that limits model interpretability and robustness. To address these challenges, we propose a Mixture-of-Experts-guided Co-Optimization (\textit{MoeCo}) framework powered by knowledge-driven learning. Within the co-optimization pipeline, to first mitigate component-level conflicts, we introduce a component-tailored adapter that disentangles task-specific features across spatial-temporal regimes, facilitating effective component specialization. Next, we develop a coordinated gradient learning strategy to handle category-level conflicts, which adaptively rebalances positive-negative gradients to enhance the perception of rare categories. Notably, inspired by surgical domain expertise, we introduce a knowledge-driven mixture-of-experts mechanism that dynamically integrates multimodal large language model-guided knowledge via activated experts, thereby enriching the co-optimization pipeline with more expressive and robust representations. Extensive experiments on the public CholecT45 and CholecT50 datasets confirm the effectiveness of the proposed co-optimization pipeline and the superiority of dynamic priors integration via the knowledge-driven mixture-of-experts mechanism. {Code will be available at \url{https://github.com/YIYIZH/MoeCo}.}

\end{abstract}

\section{Introduction}
\label{sec:introduction}
\begin{figure*}[t]
\centering
\includegraphics[width=1\textwidth]{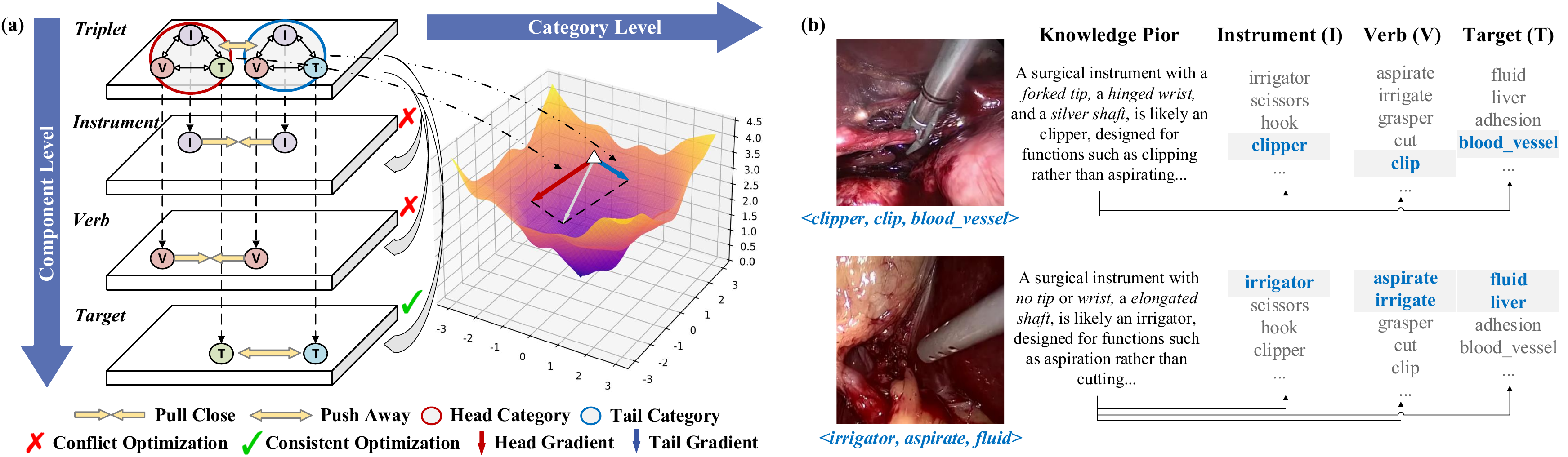}
\caption{{\textbf{(a) Hierarchical optimization conflicts in triplet (I/V/T) recognition:} We illustrate the conflict using a common head triplet (\protect\redcircle$<$\textit{grasper, retract, gallbladder}$>$) and a rare tail triplet (\protect\bluecircle$<$\textit{grasper, retract, gut}$>$). \textit{At the Component Level (Left):} In the Instrument and Verb layers, the two features are \textbf{pulled closer} because they share the same instrument (\textit{grasper}) and verb (\textit{retract}). However, in the Target and Triplet layer, they must be \textbf{pushed apart} to distinguish \textit{gallbladder} from \textit{gut} and the two triplet categories. \textit{At the Category Level (Right):} This conflict is exacerbated by dataset imbalance. The gradient from the head category (red arrow) \textbf{dominates} the optimization direction, overpowering the tail category (blue arrow). \textbf{(b) Domain Priors:} Case studies show how integrating explicit knowledge helps the reasoning process of valid triplets. For example, knowing a \textit{forked tip} tool is likely a \textit{clipper} (not an \textit{irrigator}) narrows the search space, guiding the model toward valid triplets despite noisy visual cues.}}
\label{img:teaser}
\end{figure*}

\IEEEPARstart{S}{urgical} activity understanding \cite{jin2021temporal,guo2025surgical,yang2025surgpetl} that provides context-aware decision support is a fundamental enabler in designing surgical robotic assistance systems \cite{dagnino2024robot,zheng2025survey}, offering the potential to improve surgical precision, reduce tissue trauma, and accelerate recovery times \cite{moglia2021systematic}. Particularly, surgical action triplet recognition \cite{nwoye2020recognition} aims to identify the essential components of surgical procedures and classify their interactions. It is typically formalized as a series of triplets $<$\textit{instrument, verb, target}$>$, which considers the complex relationship between surgical instruments and target tissues. This task enables understanding physician-patient interactions at a fine-grained level and provides comprehensive surgical procedural information \cite{nwoye2023cholectriplet2021}. 
However, surgical triplet recognition poses a unique challenge, as it requires the simultaneous and accurate identification of three components, where an error in any one results in an incorrect overall outcome.

Mainstream methods \cite{li2023mt, gui2024tail,peiinstrument} for surgical triplet recognition typically adopt a multi-task learning paradigm, incorporating three distinct auxiliary tasks: instrument, verb, and target recognition. For instance, MT-FiST \cite{li2023mt} introduced a multi-task fine-grained spatial-temporal framework that leverages a multi-label mutual channel loss to enhance inter-task coordination. Recognizing the significance of instruments in surgical practice \cite{nwoye2022rendezvous}, existing methods \cite{liu2024surgical, sharma2023rendezvous} have prevalently adopted weakly-supervised signals from instruments, such as class activation maps \cite{jung2021towards} or bounding boxes, to identify instruments and enhance verb and target recognition. However, current methods rely on coarse localization cues, neglecting the fine-grained structural patterns and domain priors inherent in surgical instruments. Additionally, misaligned inter-task objectives among tasks or intra-task ones across categories in triplet recognition often lead to conflicting gradients that hinder optimization \cite{ban2024fair}. As a result, despite progress in the multi-task learning paradigm, unresolved hierarchical optimization conflicts and the lack of domain knowledge integration remain significant challenges for this domain-intensive task.

Specifically, \textbf{at the component level}, as illustrated in Fig.~\ref{img:teaser} (a), decomposed tasks often result in conflicting inter-task optimization objectives \cite{ban2024fair}. For example, consider two frames belonging to different triplet categories, such as {\textit{$<$grasper, retract, gallbladder$>$} and \textit{$<$grasper, retract, gut$>$}}. For instrument and verb recognition tasks, these frames should be pulled closer in the feature space due to their shared instrument-verb association. However, the same frames need to be pushed apart for target and triplet recognition tasks to distinguish between gallbladder and gut as distinct targets. Such ambiguous objectives, derived from an entangled feature space, introduce instability in representation learning and necessitate task-specific feature adaptation. 
Within the triplet task, the uneven distribution of categories exacerbates the optimization conflicts \textbf{at the category level}. Surgical procedures inherently involve an imbalanced frequency of triplet categories due to the intricate nature of surgeons manipulating instruments on tissues \cite{peiinstrument}. This imbalance leads to the dominance of gradients from head classes, causing them to converge quickly, while rare categories remain poorly optimized. Thereby, most models fail to recognize rare but clinically significant triplet categories.

Furthermore, in the triplet recognition task as illustrated in Fig. \ref{img:teaser} (b), \textbf{domain priors} exist in the form of structural–functional constraints that define valid combinations. These priors stem from the physical design and intended use of instruments. For example, a \textit{clipper} with a \textit{forked tip} is suited for \textit{clipping} solid tissues like \textit{vessels}, making associations such as \textit{$<$clipper, aspirate, fluid$>$} implausible. Such priors reflect consistent patterns in surgical practice, where an instrument’s structure not only aids in the identification of the instrument itself, but also inherently restricts feasible actions and applicable targets. Incorporating these priors into the model introduces high-level semantic dependencies that improve recognition accuracy and robustness, particularly under challenging conditions such as motion blur, blood occlusion, and smoke interference ~\cite{luo2017vision,pei2025restore}, where visual cues alone are unreliable. However, existing methods often overlook these structural dependencies, limiting their ability to fully leverage the domain knowledge embedded in surgical contexts. 

Based on the above observations, we propose a novel Mixture-of-Experts-guided Co-Optimization (\textit{MoeCo}) framework for surgical triplet recognition, which co-optimizes hierarchical conflicts at component and category levels while employing an innovative mixture-of-experts mechanism for dynamic knowledge integration. Concretely, a component-tailored adapter (CTA) is used to create specialized representations via task-specific prompt tuning to mitigate component-level conflicts. However, standard prompt tuning implicitly assumes that both temporal and spatial features can be modeled within a shared subspace, which fails to meet the requirement for frame-wise semantic fidelity and long-range temporal consistency in surgical triplet recognition. To overcome this limitation, we distribute computation across separate latent spatial and temporal regimes to enable comprehensive feature specialization. At the category level, we propose a coordinated gradient learning (CGL) strategy to improve the perception of rare categories. This strategy rebalances the positive-negative gradients between head and tail classes, fostering a more coordinated learning process and mitigating gradient conflicts across categories. Furthermore, we leverage multimodal large language models (MLLMs) \cite{hurst2024gpt} for instrument structural knowledge mining. Drawing inspiration from the strong capacity of mixture-of-experts frameworks to harness specialized expertise, we introduce an innovative mixture-of-experts mechanism that explicitly integrates structural priors as experts, thereby distilling domain knowledge into representation modeling. The main contributions of this paper are as follows:

\begin{itemize}
\item We introduce \textit{MoeCo}, a novel framework that tackles hierarchical optimization conflicts at both the component and category levels, ensuring seamless integration of domain knowledge.
\item  Within the co-optimization pipeline, we propose a component-tailored adapter that enables feature specialization across spatial-temporal regimes to mitigate component-level conflicts. Meanwhile, a coordinated gradient learning strategy is adopted to handle category-level conflicts by balancing positive-negative gradients.
\item We enrich the co-optimization pipeline with more expressive representations via a knowledge-driven mixture-of-experts mechanism, which leverages MLLMs for knowledge mining and incorporates an innovative mixture-of-experts mechanism for dynamic knowledge integration.
\item Extensive experiments on the CholecT45 and CholecT50 datasets \cite{nwoye2022data} demonstrate that \textit{MoeCo} achieves state-of-the-art performance in surgical triplet recognition.
\end{itemize}
\section{Related Work}
\label{rw}
\subsection{Surgical Triplet Recognition}
Surgical triplet recognition, which involves identifying instrument, verb, target, and their combinations, is a complex multi-task learning problem plagued by a long-tailed data distribution \cite{nwoye2020recognition}. The mainstream multi-task learning paradigm benefiting from cross-task collaborative promotion has shown promising performance in identifying surgical action triplets \cite{li2023mt}. For instance, Li \textit{et al}. \cite{li2024surgical} developed a multi-task prior-reinforced and cross-sample network to improve synergy between component and triplet recognition. A key technique within this paradigm involves leveraging weakly-supervised signals from instruments to enhance verb and target recognition \cite{liu2024surgical}. For example, RDV \cite{nwoye2022rendezvous} employed a class activation-guided attention mechanism to extract instrument-related activations, which were then used to improve verb and target recognition. Similarly, Liu \textit{et al}. \cite{liu2024surgical} utilized the medical foundation model MedSAM to create pseudo-labels for instrument localization, providing auxiliary supervision. However, these methods often exploit only coarse localization cues, overlooking the fine-grained structural patterns within instruments. Furthermore, sub-tasks in triplet recognition may have misaligned objectives, causing their gradients to pull shared model parameters in opposing directions \cite{ban2024fair}. Consequently, despite the demonstrated potential of multi-task frameworks, they typically overlook the optimization conflicts and lack domain knowledge integration, leading to sub-optimal performance in surgical triplet recognition.

Another major challenge in surgical triplet recognition is the class imbalance problem, where models are biased toward well-represented head classes, resulting in poor generalization for underrepresented tail classes. Existing solutions for class imbalance, such as re-sampling techniques \cite{estabrooks2004multiple} and re-weighting methods \cite{zheng2020deep}, have been widely applied in general datasets. However, while these methods \cite{lin2024distributionally} have shown promise in general domains, their performance gains are marginal in specialized domains like surgery, where the imbalance is far more severe. For example, in the widely used CholecT45 dataset \cite{nwoye2022data}, the most frequent triplet category contains over 40,000 samples, whereas the least frequent class has only 8 samples. To tackle this challenge, MT4MTL-KD \cite{gui2024mt4mtl} employed teacher models trained on less imbalanced sub-tasks, such as instrument recognition, to guide a student model handling the more complex and imbalanced triplet recognition task. Recently, TERL \cite{gui2024tail} introduced a memory bank for contrastive learning, specifically designed to improve tail-class recognition. However, these approaches mainly focus on the instance-level rebalancing, neglecting the gradient-level optimization conflict, which is crucial for balanced training across categories.

\subsection{Mixture-of-Experts Framework}
The mixture-of-experts (MoE) framework \cite{jacobs1991adaptive} has emerged as a powerful paradigm for leveraging task-specific expertise in various machine learning domains, including natural language processing, computer vision, and multi-task learning. MoE aims to decompose complex tasks into sub-components, with specialized experts trained to handle specific parts of the problem. Shazeer \textit{et al}. \cite{shazeer2017outrageously} first proposed the sparsely gated MoE, which uses a sparse gating mechanism to activate only a subset of experts for each input, dramatically reducing computation costs. Later, Guo \textit{et al}. \cite{guo2018dynamic} introduced the dynamic task prioritization network, using a gating mechanism to dynamically allocate resources to different tasks based on their difficulty. With the advent of Transformer architectures, this concept was further expanded in Transformer-based models such as switch Transformers \cite{fedus2022switch}, which scale to trillions of parameters while maintaining high computational efficiency. Unlike conventional MoE frameworks that often rely on randomly initialized learnable experts, our approach leverages predefined knowledge-driven experts tailored specifically to semantic patterns extracted from surgical contexts. This ensures that each expert is specialized in handling unique structural or semantic components of surgical instruments, providing a more meaningful and task-specific initialization. 

\section{Methodology}
\begin{figure*}[t]
\includegraphics[width=\textwidth]{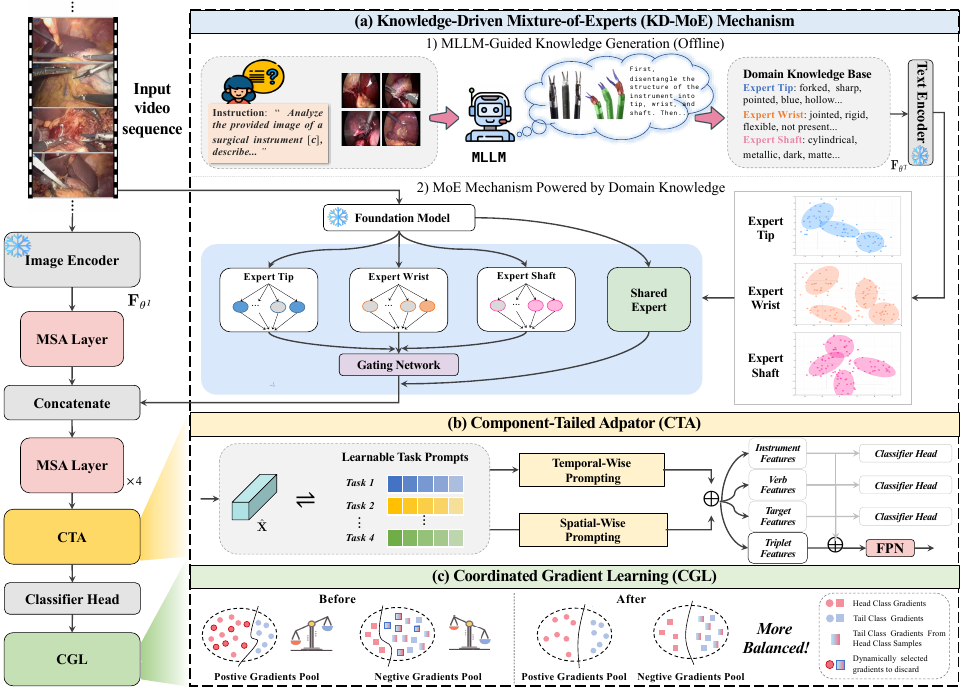}\caption{
Overview of the proposed Mixture-of-Experts-guided Co-optimization (\textit{MoeCo}) framework. We first employ an MLLM to generate \textbf{domain knowledge} used in constructing Gaussian models for each expert. During training, the original visual embeddings from an input video sequence are enriched by concatenating them with MoE-purified semantics. Multi-head self-attention layers process the enriched embeddings, followed by a \textbf{component-tailored adapter} for task-specific feature modeling via temporal-spatial prompting. Individual component features drive their recognition outputs, while triplet features integrate with them and undergo refinement with a feature pyramid network. Finally, the \textbf{coordinated gradient learning} strategy is employed to decompose and rebalance positive-negative gradients for better perception of rare triplet categories.}
\label{img:framework}
\end{figure*}

\subsection{Overview}
The proposed mixture-of-experts-guided co-optimization framework \textit{MoeCo}, illustrated in Fig. \ref{img:framework}, comprises three pivotal modules: a knowledge-driven MoE (KD-MoE) mechanism, a component-tailored adapter (CTA), and a coordinated gradient learning (CGL) strategy. To distill domain knowledge into feature modeling, the KD-MoE mechanism (Sec. \ref{sec:moe}) captures instrument-anchored dependencies by integrating relative knowledge patterns through activated experts. At the component level, CTA  (Sec. \ref{sec:cta}) is adopted to model task-specific representations, mitigating conflicting optimization objectives arising from a shared latent space. To tackle category-level conflicts, the CGL strategy  (Sec. \ref{sec:cgl}) is proposed for enhanced perception of rare categories by rebalancing positive-negative gradients during training.

\subsection{Knowledge-Driven Mixture-of-Experts Mechanism}
\label{sec:moe}

The essence of surgical action triplet recognition lies in the deep understanding and description of surgical scenes. Inspired by domain priors in surgical practice, the structural design of surgical instruments inherently encapsulates rich semantic patterns, which are crucial not only for accurate instrument recognition but also for reliably identifying associated verbs and targets. Instinctively, we recognize that incorporating semantic knowledge from text-based priors can facilitate the semantic perception of triplet models. To reach this purpose, we employ an MLLM (e.g., GPT-4o \cite{hurst2024gpt}) to achieve the efficient extraction of fine-grained structure-specific knowledge given target images. Leveraging the strong generalizability of pretrained visual-language models \cite{radford2021learning, peiinstrument}, we convert domain knowledge into phrase-based textual embeddings. To distill the textual prior knowledge into the model, we propose the knowledge-driven mixture-of-experts mechanism that enables seamless integration of relevant semantics.

\subsubsection{MLLM-guided Knowledge Generation}
Surgical instruments can be dissected into three core fragments, with each isolated structure showcasing diverse visual characteristics. To build a comprehensive semantic foundation of fine-grained expert knowledge, we employ a systematic prompting strategy to query the MLLM. For each surgical instrument class $c$ from a dataset, we present the MLLM with a set of representative images $I_c$ ($\geq3$) of that instrument for richer responses. {The selection process of the representative images is guided by two requirements: 
(1) {Visual Clarity:} The key structures of the instrument must be clearly visible, with minimal interference from occlusion, motion blur, smoke, or blood. 
(2) {Comprehensive Coverage:} The selected set of images for a given instrument must collectively capture all critical components, namely the tip, wrist, and shaft, from informative viewpoints. In practice, we used three images for each instrument class by default, except for the ``grasper'', for which we used four images because an additional view was needed to clearly reveal its hollow tip structure.}

These images are accompanied by a crafted text prompt designed to elicit disentangled descriptions of the instrument's key functional parts, including candidate attributes of tip $\{t_c\}$, wrist $\{w_c\}$, and shaft $\{s_c\}$. {To prevent hallucinated or erroneous attributes from corrupting the feature space, we implement a rigorous human-in-the-loop safeguard. Specifically, raw candidate attributes undergo strict pruning to eliminate redundant, generic, or visually non-discriminative terms. Furthermore, when a specific structure is absent from an instrument, we explicitly assign a ``not present'' indicator to capture structural existence. This safeguard procedure is executed by three researchers using a consensus-building approach, which effectively reduces the average number of attributes per class from 11.33 to 6.67. Finally, the retained attributes are compiled into a formalized knowledge base.}

This extraction process is repeated for all instrument classes $\mathcal{C}$, resulting in a comprehensive repository of three decoupled knowledge sets $\mathcal{\{T, W, S\}}$, where $\mathcal{T} = \bigcup_{c \in \mathcal{C}} \{t_c\}$, $\mathcal{W} = \bigcup_{c \in \mathcal{C}} \{w_c\}$ and $\mathcal{S} = \bigcup_{c \in \mathcal{C}} \{s_c\}$. Taking the characteristic description ``forked'' in the knowledge set $\mathcal{T}$ as an example, we then adopt predefined templates such as ``The tip of the surgical instrument is \{forked\}'' to improve the compatibility of decoupled knowledge with the text encoder. This can improve the compatibility of decoupled knowledge with the text encoder (e.g., CLIP \cite{radford2021learning}), ensuring better alignment between textual and visual representations.
We also consider the case where there is a background class while no instrument occurs in the image. With ``not present'' as descriptions added in both tip, wrist, and shaft knowledge sets, we can implicitly enhance the feature with additional information about the presence of an instrument. Finally, the formulated textual sentences from the knowledge base are converted into high-dimensional vectors by a pre-trained text encoder $\mathbf{F}_{\theta^T}$.

\subsubsection{MoE Mechanism Powered by Domain Knowledge} Current MoE methodologies typically rely on randomly initialized, black-box sub-networks, which lack interpretability and hinder the seamless integration of expert knowledge. Unlike these mainstream MoE mechanisms \cite{shazeer2017outrageously}, our approach instills domain knowledge into feature modeling by leveraging three experts predefined by semantic priors of tip, shaft and wrist. This ensures that each expert specializes in distinct structural components of surgical instruments, providing more meaningful and task-specific initialization. Specifically, we propose a knowledge-driven MoE mechanism where each distinct expert $\{\mathcal{E_{{T}}, E_{{W}}, E_{{S}}}\}$ is initialized as a mixture of Gaussian distributions \cite{najar2017comparison}. 

{Taking the tip expert $\mathcal{E}_{\mathcal{T}}$ as an example, each retained tip attribute is treated as one Gaussian component. The MLLM-guided knowledge base is constructed at the instrument-category level, recording the attributes associated with each instrument class. Therefore, the assignment of image features to Gaussian components is performed offline using a hard attribute-conditioned strategy. Specifically, for the $j$-th attribute $a_{\mathcal{T}}^j$, we collect the normalized image features from training images whose instrument category contains this attribute in the MLLM-derived knowledge base. These features are assigned to the corresponding Gaussian component. If multiple attributes are associated with the same instrument category, the image features from this category can contribute to multiple relevant Gaussian components, since different visual attributes may coexist in the same instrument.} The MLLM-guided knowledge sets and Gaussian models are established offline before the formal training process begins, with MLLM unused in both training and inference stages.

During training, we utilize a pretrained visual encoder of a visual-language model to extract each input image embeddings $\mathbf{x}_f$. Given $\mathbf{x}_f$, we conduct probabilistic pooling to compute the likelihood values from each expert knowledge set. Specifically, considering the expert $\mathcal{E_{{T}}}$ with $\mathcal{J}$ attributes, the probability density of a feature vector $\mathbf{x}_f$ is given by:
\begin{equation}
\mathcal{E_{{T}}}(\mathbf{x}_f)= \bigcup_{j \in \mathcal{J}} \left\{\mathcal{N}(\mathbf{x}_f|\mu_\mathcal{T}^{j},\Sigma_\mathcal{T}^{j})\right\}.
\end{equation}
Here, $ \mu_\mathcal{T}^{j}$ and $\Sigma_\mathcal{T}^{j}$ represent the mean vector and covariance matrix of the $j$-th Gaussian component, respectively. {The mean vector $ \mu_\mathcal{T}^{j}$ is initialized using the normalized textual representation of the attribute, i.e., $ \mathbf{F}_{\theta^T}(a_{\mathcal{T}}^j)$. The covariance matrix $\Sigma_\mathcal{T}^{j}$ is estimated by:
\begin{equation}
\Sigma_\mathcal{T}^{j} =\operatorname{diag}
\left(
\frac{1}{|\mathcal{X}_{\mathcal{T}}^j|}
\sum_{\bar{{x}}\in\mathcal{X}_{\mathcal{T}}^j}
(\bar{{x}}-\mu_{\mathcal{T}}^{j})^2
\right),
\end{equation}
where $\mathcal{X}_{\mathcal{T}}^j$ denotes the set of normalized image features associated with attribute $a_{\mathcal{T}}^j$, $\bar{{x}}={x}/\|{x}\|_2$ is the normalized image feature, and the square operation is applied element-wise.}
This process is repeated for the other experts, $\mathcal{E_{{W}}}$ and $\mathcal{E_{{S}}}$, ensuring that each expert is tailored to its respective attributes. To identify the most relevant attributes, we employ a gating network that activates the top-$k$ attributes from each expert based on their probabilistic values. The gating function is defined as:
\begin{equation}
\label{equ:topk}
\mathcal{G}(\mathbf{x}_f) = \left\{ \mu_i^{(j)} \ \middle| \ i \in \mathcal{\{T,W,S}\},\  j \in \operatorname{TopK}\left(\mathcal{E}_i(\mathbf{x}_f), k\right) \right\}
\end{equation}

By capturing the inherent distribution of each attribute within the dataset, the proposed knowledge-driven MoE mechanism dynamically activates the most relevant semantic knowledge into the representation, enabling the model to adapt effectively to the structural attributes of the input images.

{To further enhance the representation, we incorporate a learnable multi-head self-attention (MSA) layer with default PyTorch random initialization that serves as a shared expert \cite{dai2024deepseekmoe}, consistently activated to capture and consolidate common knowledge across diverse contexts.} By compressing common knowledge into the shared expert, this layer captures generalizable information overlooked by specialized experts, thereby fully leveraging the comprehensive visual representations of pretrained visual-language models. The outputs from the shared expert $\mathcal{E}_{shared}$ and the specialized experts are concatenated as follows:
\begin{equation}
\mathbf{x}_{moe} = \operatorname{Concat}\left(\mathcal{E}_{shared}(\mathbf{x}_f), \mathcal{G}(\mathbf{x}_f)\right),
\end{equation}
where $\mathbf{x}_{moe}$ represents the combined knowledge from both shared and specialized experts. This enhanced representation is then concatenated with the original visual feature vector.

\subsection{Component-Tailored  Adapter} 

Previous methods such as TERL \cite{gui2024tail} have constructed task-specific branches with distinct classifiers but without modeling distinct features for each task, resulting in unstable training due to optimization competition. In contrast to common strategies that seamlessly introduce additional parallel networks to capture task-specific features, prompt learning \cite{zhou2022learning} offers an alternative approach by utilizing task-specific learnable prompts for feature specialization in a parameter-efficient way. However, standard spatial-wise prompt tuning only captures a short-term view that is insufficient for triplet action identification. Given the intricacy of the entire surgical procedure, triplet action components are difficult to recognize due to challenges such as motion blur, blood coverage, and smoke interference in real scenarios \cite{luo2017vision}. Effectively relating the information that is adjacent in time to what is happening in the present can benefit accurate predictions on the current frame. Therefore, we advocate for the integration of spatial and temporal prompting to ensure both frame-wise semantic fidelity and long-range temporal consistency in analyzing surgical videos.
\label{sec:cta}
\begin{figure}[t]
\centering
\includegraphics[width=8.5cm]{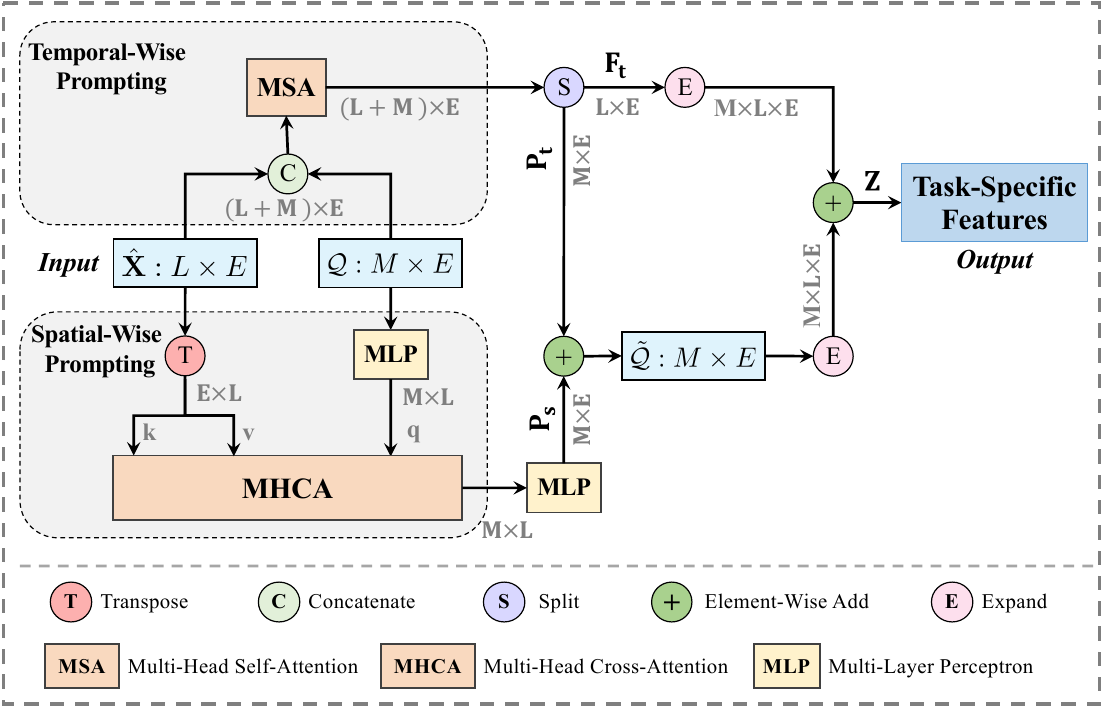}
\caption{An illustration of the component-tailored adapter. CTA learns the temporal- and spatial-wise task-specific features in a parameter-efficient way with no need for multiple network architectures for each sub-task.}
\label{img:stp}
\end{figure}
As illustrated in Fig. \ref{img:stp}, distinct task prompts are embedded with the task-agnostic video features to operate attention mechanisms from both spatial and temporal dimensions for more comprehensive feature interaction. We first randomly initialize a small set of dedicated, learnable prompt tokens $\mathcal{Q} \in \mathbb{R}^{M \times E}$, where $M$ is the number of tasks and $E$ is the dimension of each prompt embedding. Given the video-wise feature $\hat{\mathbf{X}} \in \mathbb{R}^{L \times E}$, where $L$ represents the sequence length, we first conduct temporal-wise prompting. Specifically, we concatenate the video embedding and prompt tokens and feed them into an MSA layer:
\begin{equation}
\mathbf{F}_{t},\mathbf{P}_{t}=\operatorname{Split}(\operatorname{MSA}(\operatorname{Concat}(\hat{\mathbf{X}},\mathcal{Q}))),
\end{equation}
where $\mathbf{F}_{t}\in\mathbb{R}^{L\times E}$ is the basic temporally-enhanced feature map and $\mathbf{P}_{t}\in\mathbb{R}^{M\times E}$ represents the temporally-enhanced prompt tokens. Concurrently, to capture spatial information, we employ a multi-head cross-attention (MHCA) mechanism. The learnable prompts $\mathcal{Q}$ are transformed by a multilayer perceptron (MLP) to form the query $q \in \mathbb{R}^{M \times L}$, while the transposed video-wise feature $\hat{\mathbf{X}}^\top \in \mathbb{R}^{E \times L}$ serves as the key $k$ and value $v$. The resulting spatially aware prompts $\mathbf{P}_{s}\in\mathbb{R}^{M\times E}$ are computed by:
\begin{equation}
\mathbf{P}_{s}=\operatorname{MLP}(\operatorname{MHCA}(\operatorname{MLP}(\mathcal{Q}),\hat{\mathbf{X}}^\top,\hat{\mathbf{X}}^\top)).
\end{equation}

Finally, the temporal and spatial prompts are integrated through element-wise addition to form the final task-specific prompts $\tilde{\mathcal{Q}}=\mathbf{P}_{t}+\mathbf{P}_{s}$. To generate the task-specific features, we expand the dimensions of both the base feature embeddings and the refined prompts for seamless summation:
\begin{equation}
\mathbf{Z}=\operatorname{Expand}(\mathbf{F}_t)+\operatorname{Expand}(\tilde{\mathcal{Q}})\in\mathbb{R}^{M\times L\times E},
\end{equation}
where $\mathbf{Z} = \{\bm{z_i}, \bm{z_v}, \bm{z_t}, \bm{z_{ivt}}\} $ and $M = 4$. {The indices $i$, $v$, $t$, and $ivt$ denote the instrument, verb, target, and triplet branches, respectively.} The final triplet feature is weighted fused by a hyperparameter $\alpha$, denoted as:
\begin{equation}
\label{eq:zivt}
\bm{z_{ivt}}= \alpha(\bm{z_i}+ \bm{z_v}+ \bm{z_t})+ \bm{z_{ivt}}.
\end{equation}

\subsection{Coordinated Gradient Learning}
\label{sec:cgl}
Due to the severe long-tailed data distribution in surgical triplet datasets, e.g., CholecT45, existing loss functions struggle to handle this imbalance effectively. Specifically, we observe a pronounced issue of inconsistent optimization flow, where head classes, due to their abundance, converge quickly and begin to overfit, while tail categories remain underfitted throughout the training process. This imbalance leads to asynchronous learning dynamics, resulting in suboptimal performance. To better investigate the detailed optimization procedure, we dissect the vanilla binary cross-entropy (BCE) loss into positive $\mathcal{L}^+$ and negative $\mathcal{L}^-$ components \cite{tan2020equalization}. Let $\mathcal{O} = \{(\mathbf{x}_n, {y}_n)\}_{n=1}^\mathcal{N}$ be a batch of data where $\mathbf{x}_n$ is an input instance and ${y}_n \in \{0,1\}^\mathcal{D}$ is its corresponding label vector for $\mathcal{D}$ categories. The decomposed loss can be denoted as: 
\begin{equation}
\begin{aligned}  
\mathcal{L}_{\text{BCE}} = -\sum_{n=1}^{\mathcal{N}} \sum_{d=1}^\mathcal{D} \underbrace{y_{n,d}\log\sigma(\bm{w}_d^\top\bm{z_{ivt}}^n)}_{\mathcal{L}^+} + \\
\underbrace{(1-y_{n,d})\log(1-\sigma(\bm{w}_d^\top\bm{z_{ivt}}^n))}_{\mathcal{L}^-},
\end{aligned}
\end{equation}
where $\bm{\omega}_d \in \mathbb{R}^E$ represents the classifier weights for triplet category $d$ and $\sigma(\cdot)$ is the sigmoid function. For clarity, we omit the instance index $n$ and task index $ivt$, and focus on a single triplet class $d$. Suppose the logit $z_d = {w}_d^\top\bm{z_{ivt}}^n$, the gradient of BCE loss with respect to $z_d$ is:
\begin{equation}
\frac{\partial \mathcal{L}_{BCE}}{\partial z_d} = \sigma(z_d) - y_d.
\end{equation}
For a positive sample ($y_d = 1$), the gradient is $\sigma(z_d) - 1$, and for a negative sample ($y_d = 0$), it is $\sigma(z_d)$.

\begin{figure}[t]
\centering
\includegraphics[width=9cm]{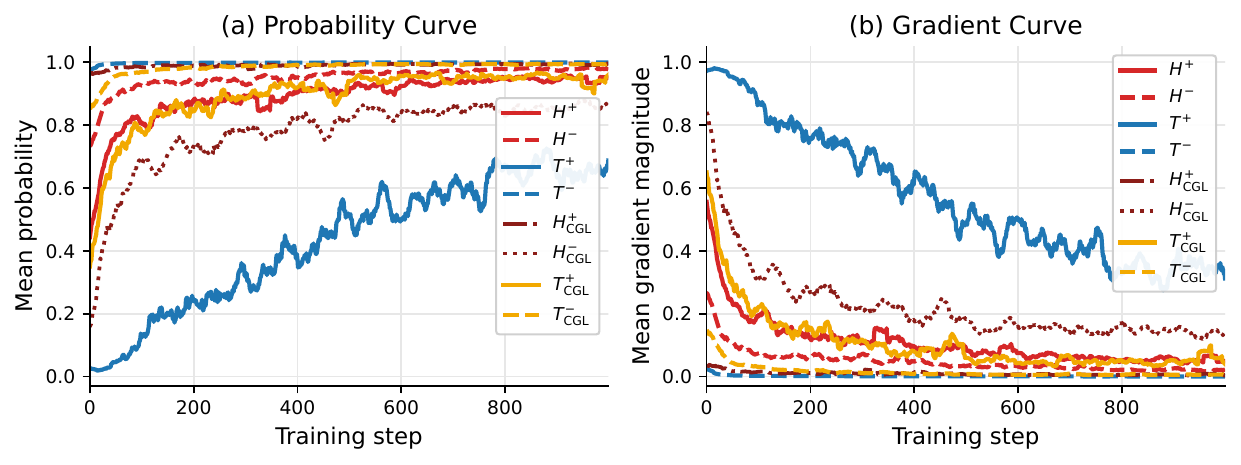}
\caption{
{Predicted mean probability and mean gradient curves on the CholecT45 dataset during training.
H and T denote head and tail classes, respectively, while \(+\) and \(-\) denote positive and negative samples.
\(H^{+}_{\mathrm{CGL}}\), \(H^{-}_{\mathrm{CGL}}\), \(T^{+}_{\mathrm{CGL}}\), and \(T^{-}_{\mathrm{CGL}}\) denote the corresponding curves after applying CGL.
The gradient magnitude is computed from the predicted probability.
Compared with BCE, CGL substantially reduces the discrepancy in optimization behavior between tail and head classes.}
}
\label{img:grad}
\end{figure}
\begin{table*}[t]
\caption{Quantitative comparisons between the proposed method and the SOTA methods using 5-fold cross-validation on the CholecT45 dataset. The mean and standard deviation results of AP are reported. \textbf{Best} and \underline{Second Best}.}
\centering
\fontsize{9pt}{9pt}\selectfont
    % \addtocounter{table}{0}
    \renewcommand{\arraystretch}{1.2}
    \setlength{\tabcolsep}{3.5mm}{\resizebox{1\linewidth}{!}{
\begin{tabular}{@{}lccccccc@{}}
\toprule
\textbf{Method }& \textbf{Backbone} & $\boldsymbol{AP_I}$ & $\boldsymbol{AP_V}$ & $\boldsymbol{AP_T}$ & $\boldsymbol{AP_{IV}}$ & $\boldsymbol{AP_{IT}}$ & $\boldsymbol{AP_{IVT}}$ \\
\midrule
TripNet \cite{nwoye2020recognition} & ResNet-18 & 89.9±1.0 & 59.9±0.9 & 37.4±1.5 & -- & -- & 24.4±4.7 \\
RDV \cite{nwoye2022rendezvous} & ResNet-18 & 89.3±2.1 & 62.0±1.3 & 40.0±1.4 & 34.0±3.3 & 30.8±2.1 & 29.4±2.8 \\
RiT \cite{sharma2023rendezvous} & ResNet-18 & 88.6±2.6 & 64.0±2.5 & 43.4±1.4 & 38.3±3.5 & 36.9±1.0 & 29.7±2.6 \\
TripDis \cite{chen2023surgical} & ResNet-50 & 91.2±1.9 & 65.3±2.8 & 43.7±1.6 & -- & -- & 33.8±2.5 \\
SelfD \cite{yamlahi2023self} & SwinB×2+SwinL & -- & -- & -- & -- & -- & 38.5±0.0 \\
MT4MTL-KD \cite{gui2024mt4mtl} & ResNet-18+SwinL & {93.9±2.0} & \textbf{73.8±2.0} & \underline{52.1±5.2} & 46.5±3.4 & 46.2±2.3 & 38.9±1.6 \\
TERL-T \cite{gui2024tail} & SwinT & 93.1±2.4 & 71.1±1.7 & 48.9±3.9 & 44.9±4.4 & 41.9±3.1 & 35.7±2.3 \\
TERL-B \cite{gui2024tail} & SwinB & 93.5±2.4 & 72.8±2.8 & 51.3±3.8 & 47.0±5.6 & 45.7±2.8 & 38.9±2.5 \\
CurConMix-T \cite{jeon2025curconmix}  & SwinT & 90.4±2.1 &67.8±1.8 &48.3±3.4 &43.3±2.9 &43.3±1.8 &37.7±2.1\\
CurConMix-B \cite{jeon2025curconmix} & SwinB & 90.9±2.0 &68.3±1.3 &49.8±3.2 &45.2±4.2 &45.1±1.1 &39.1±2.0\\
LAM-Large (Ens) \cite{li2024parameter}
& ResNet-18+ViT-L
& \underline{94.5±1.9} 
& {72.4±2.3} 
& {51.6±2.8} 
& { \textbf{48.5±3.4}} 
& {\underline{47.2±2.2}}
& {40.3±3.1} \\
\midrule
\textbf{MoeCo-T (Ours)} & SwinT & 93.8±2.1 & 72.1±1.8 & 51.1±5.1 & {47.5±5.3} & {46.6±2.2} & {40.5±3.5} \\
\textbf{MoeCo-B (Ours)} & SwinB & \underline{94.5±1.7} & \textbf{72.9±1.3}  & {51.5±5.9} & \underline{48.4±6.3} & {46.6±2.8} & \underline{41.7±3.3}  \\
\textbf{MoeCo-Ens (Ours)} & SwinT+SwinB & \textbf{94.7±2.1} & 72.5±1.2 & \textbf{52.3±5.8} & {47.8±5.9} & \textbf{47.7±2.5} & \textbf{42.6±2.6} \\
\bottomrule
\end{tabular}}
}
\label{tab:benchmark}
\end{table*}
As depicted in Fig. \ref{img:grad}, a significant discrepancy in both output probabilities and gradient magnitudes exists between positive and negative samples of tail classes under the BCE loss. In contrast, head classes exhibit a much more balanced distribution. To equalize the gradient gaps in tail classes akin to those in head classes, a straightforward yet effective method involves hastening positive sample learning and decelerating negative sample learning in tail classes. This balance is achieved by adjusting the positive-to-negative gradient ratio from two perspectives: {1) $h_d^-$: Suppress the negative loss of a tail category $d$ (with probability $\gamma$) when the current sample belongs to a head category, preventing abundant head-category samples from producing excessive negative gradients for tail categories.} 2) $h_d^+$: Discard gradients (with probability $\gamma$) of the positive loss from head classes. Slowing down positive learning in head classes offers an alternative approach to accelerate learning in tail classes, given the competitive nature of these two loss components. The CGL loss function for one input instance $\mathbf{x}_n$ is denoted as follows:
\begin{equation}
\label{eq:cgl}
    \mathcal{L}_{\text{CGL}} = -\sum_{d=1}^{\mathcal{D}} h_{d}^+y_d \log\left(\sigma({z}_{d})\right) + h_{d}^-(1-y_d) \log\left(1-\sigma({z}_{d})\right),
\end{equation}
where ${h}_d^+ = 1-\lambda E(d)$ and ${h}_d^- = 1 - \lambda F(d)$. $E(d)$ is 1 if $d$ belongs to head category and 0 otherwise. $F(d)$ is 1 if $d$ belongs to tail category and {$\mathbf{x}_n$} belongs to head class, otherwise is 0. 
$\lambda$ is a random variable with a probability of $\gamma$ to be 1 and $1-$ $\gamma$ to be 0. {CGL keeps the positive and negative losses of medium classes unchanged, and only modulates the gradients related to head and tail categories where the optimization imbalance is most severe.}

{It is worth noting that \(H^{-}_{\mathrm{CGL}}\) exhibits a moderately larger gradient than \(H^{-}\) under BCE as shown in Fig. \ref{img:grad}. Since CGL redistributes the optimization pressure by reducing the dominance of head categories, it slightly relaxes the over-confident head-negative suppression. Meanwhile, \(H^{+}_{\mathrm{CGL}}\) remains highly confident with a low gradient magnitude, showing that positive head-class learning is preserved. The major improvement still occurs in tail classes, where \(T^{+}_{\mathrm{CGL}}\) becomes much closer to the head-class behavior.} To leverage these advantages and address the severe long-tailed data distribution at the triplet level, the overall loss function for surgical triplet recognition combines the CGL loss for the triplet task with individual BCE losses for instrument, verb, and target recognition.

\begin{figure*}[t]
    \centering
    \begin{subfigure}{0.245\textwidth}
        \includegraphics[width=\textwidth]{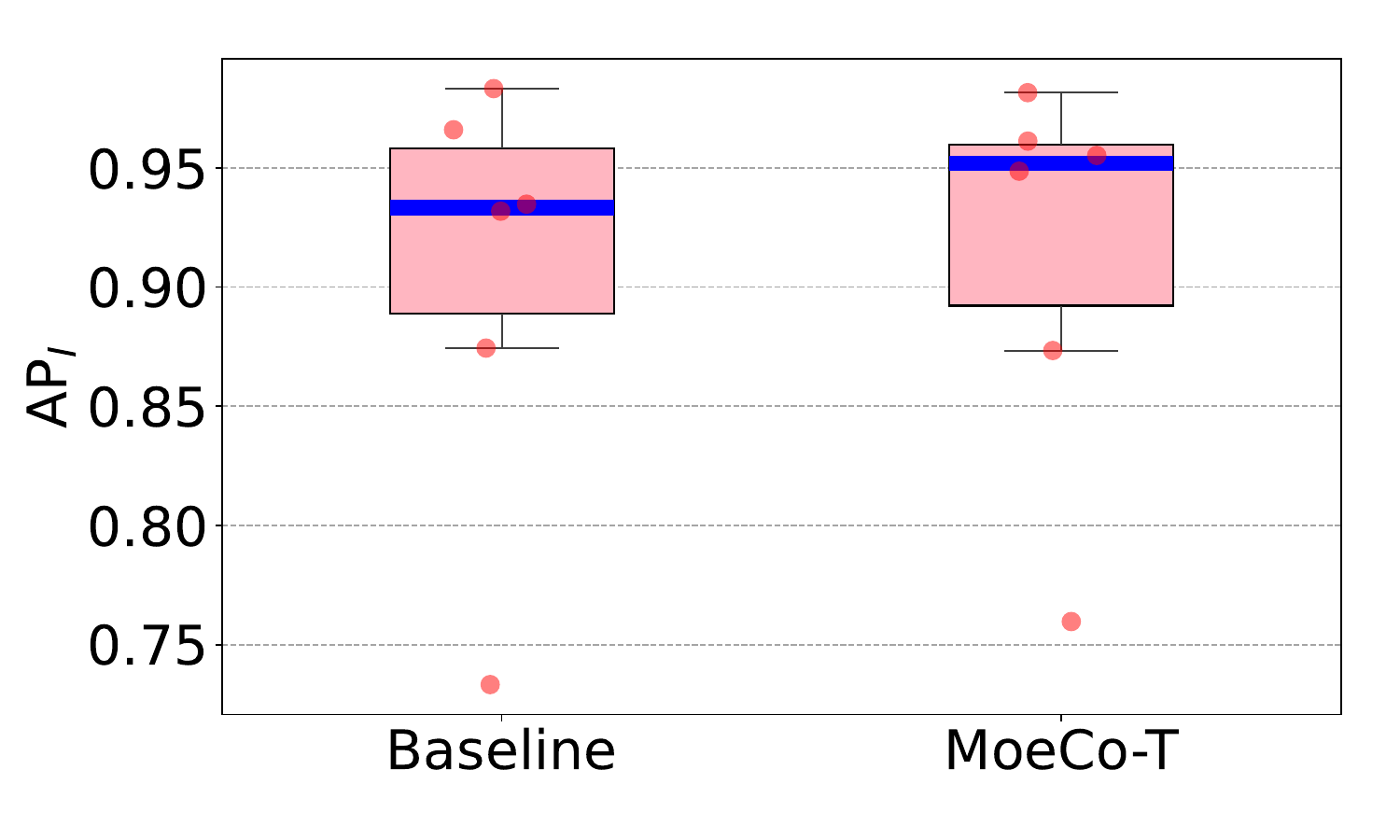}
        \caption{Task I}
    \end{subfigure}
    % \hfill
    \begin{subfigure}{0.245\textwidth}
        \includegraphics[width=\textwidth]{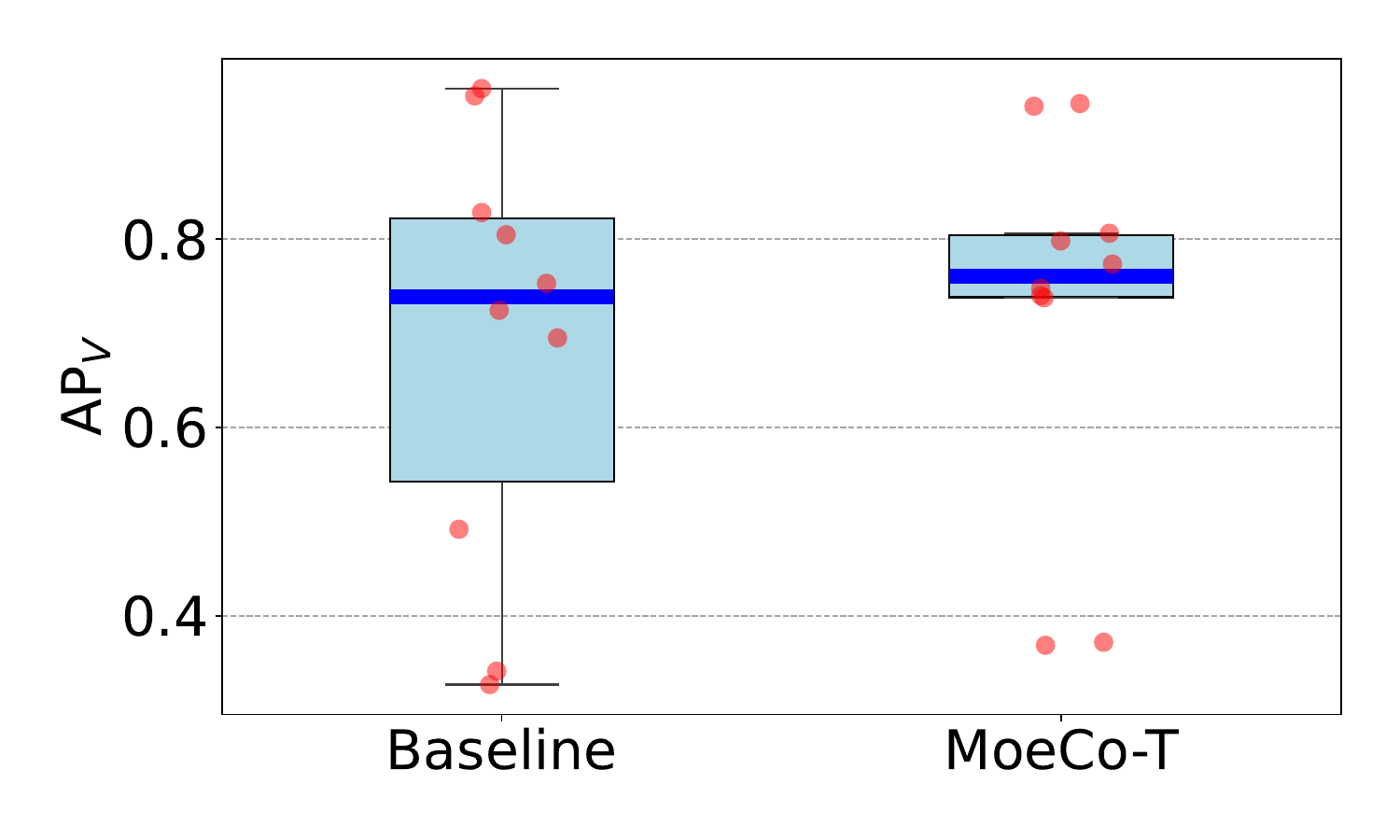}
        \caption{Task V}
    \end{subfigure}
    % \hfill
    \begin{subfigure}{0.245\textwidth}
        \includegraphics[width=\textwidth]{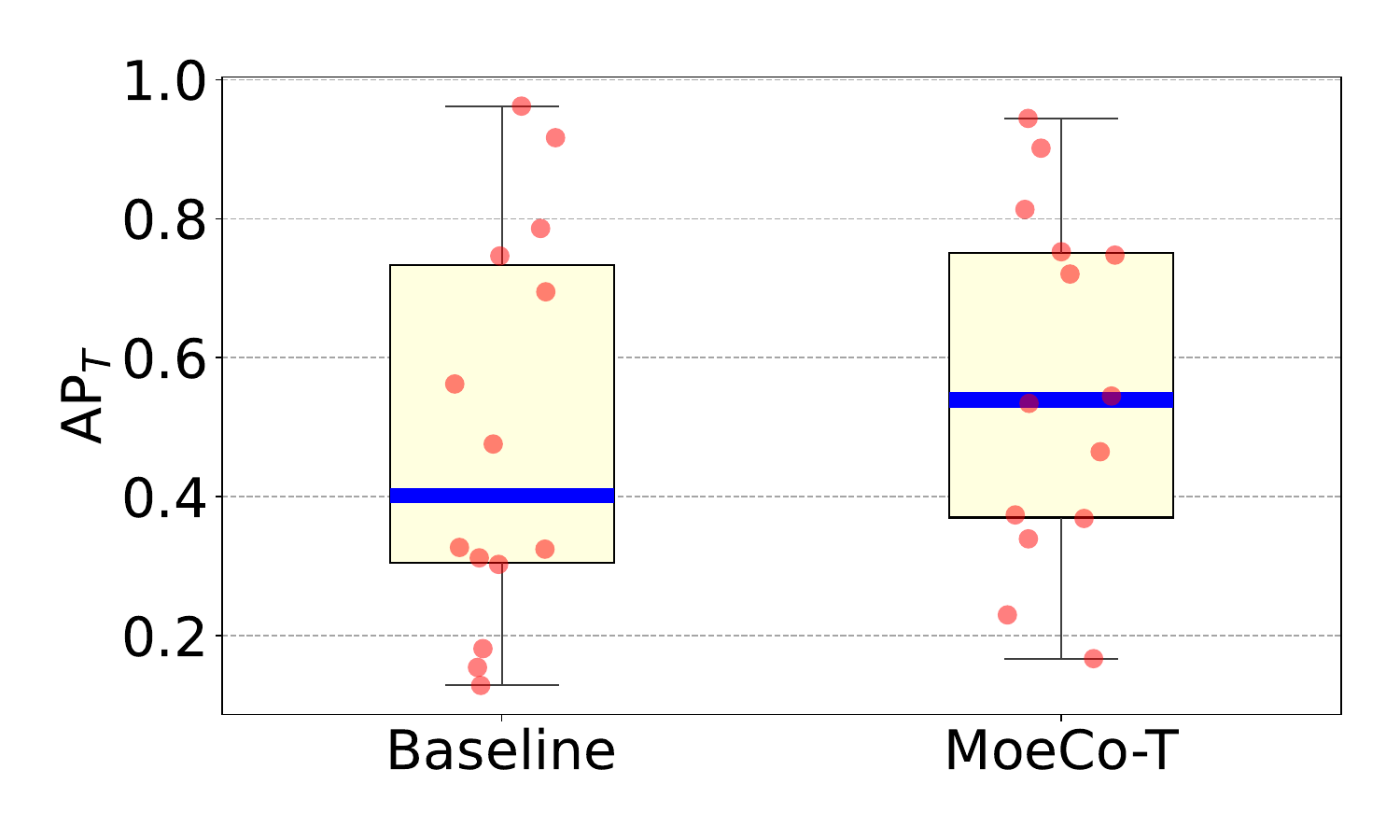}
        \caption{Task T}
    \end{subfigure}
    % \hfill
    \begin{subfigure}{0.245\textwidth}
        \includegraphics[width=\textwidth]{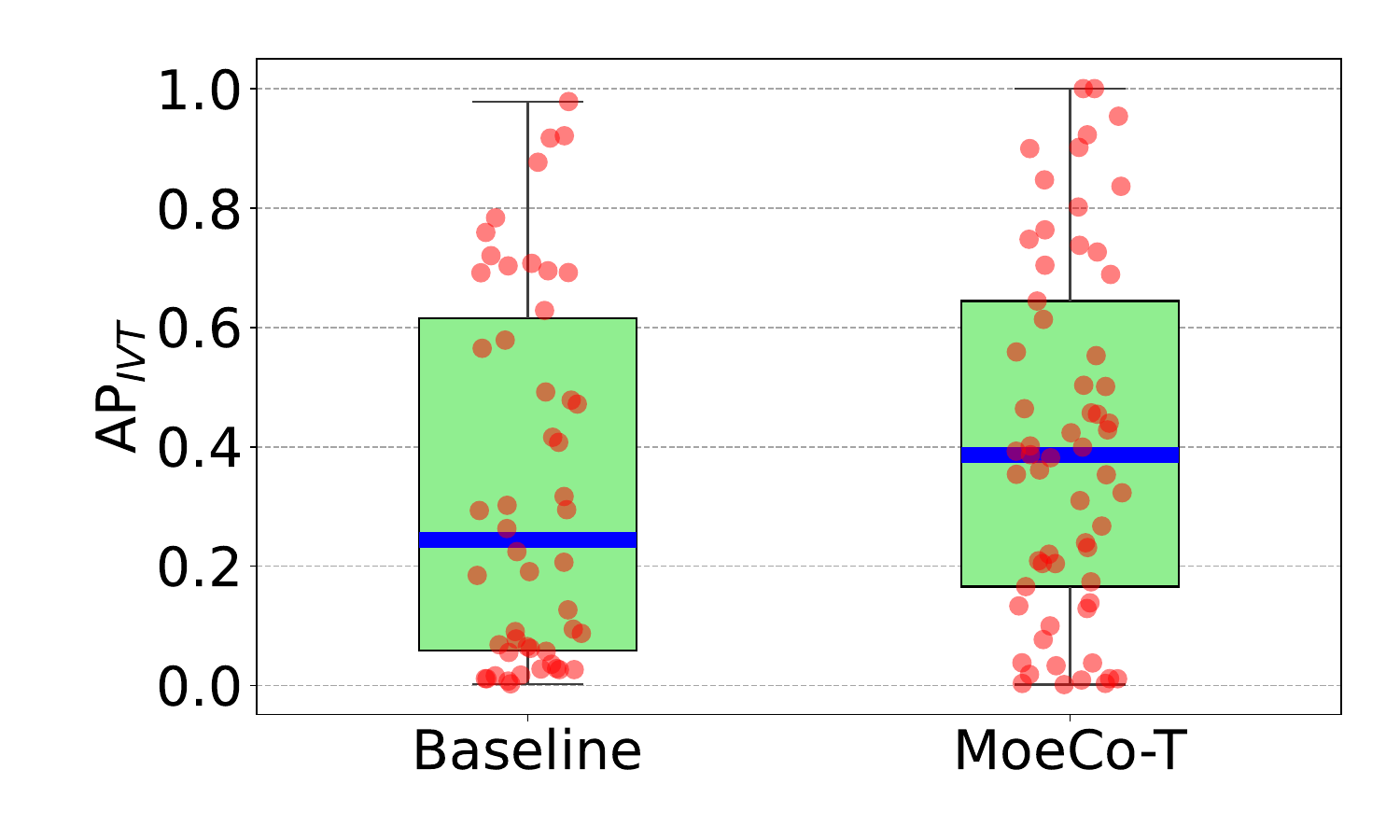}
        \caption{Task IVT}
    \end{subfigure}
    \caption{Box plots with overlaid dot distributions compare the per-category performance of the baseline and MoeCo-T across four task metrics. Each red dot represents a category, while the blue lines indicate the median AP values.}
    \label{fig:box}
\end{figure*}

% \begin{figure*}[t]
%     \centering

%     \subfloat[Task I]{
%         \includegraphics[width=0.23\textwidth]{img/zhang5-1.pdf}
%         \label{fig:box_i}
%     }
%     \hfill
%     \subfloat[Task V]{
%         \includegraphics[width=0.23\textwidth]{img/zhang5-2.pdf}
%         \label{fig:box_v}
%     }
%     \hfill
%     \subfloat[Task T]{
%         \includegraphics[width=0.23\textwidth]{img/zhang5-3.pdf}
%         \label{fig:box_t}
%     }
%     \hfill
%     \subfloat[Task IVT]{
%         \includegraphics[width=0.23\textwidth]{img/zhang5-4.pdf}
%         \label{fig:box_ivt}
%     }

%     \caption{Box plots with overlaid dot distributions compare the per-category
%     performance of the baseline and MoeCo-T across four task metrics.
%     Each red dot represents a category, while the blue lines indicate
%     the median AP values.}
%     \label{fig:box}
% \end{figure*}

\section{Experiment}\label{sec:exp}
\subsection{Datasets and Implementation Details}
\subsubsection{Introduction of Datasets} We conduct experiments on two public datasets (CholecT45 and CholecT50) from the CholecTriplet2021 challenge \cite{nwoye2023cholectriplet2021}. The CholecT45 dataset \cite{nwoye2022data} contains 45 laparoscopic cholecystectomy video sequences comprising 100.9K frames annotated with 161K triplet instance labels. Each frame includes annotations of 100 binary action triplets, consisting of 6 instruments (${I}$), 10 verbs (${V}$), and 15 targets (${T}$). We adopt the official 5-fold cross-validation strategy with a 31-5-9 split for training, validation, and testing, respectively. {Following TERL \cite{gui2024tail}, we conduct ablation studies and sensitivity analyses on the validation set of Fold 1. We further evaluate the generalizability of our method on the CholecT50 dataset, which contains 50 laparoscopic cholecystectomy videos. Consistent with the data partitioning protocol in RDV \cite{nwoye2022rendezvous}, we split CholecT50 into 35 training videos, 5 validation videos, and 10 testing videos. The hyperparameters selected on CholecT45 are directly applied to CholecT50 without additional tuning.} Performance is evaluated using average precision (AP) metrics including triplet AP ($AP_{IVT}$), association AP (${AP}_{IV}$ and ${AP}_{IT}$), and component AP (${AP}_{I}$, ${AP}_{V}$, ${AP}_{T}$), where ${AP}_{IVT}$ serves as the primary metric for complete triplet recognition \cite{gui2024tail}.

\subsubsection{Implementation Details} The baseline method consists of a Swin Transformer \cite{liu2021swin} as the spatial visual encoder \cite{gui2024tail} to extract image features, and a temporal backbone \cite{zhang2022actionformer} to model video-wise interactions. The temporal backbone consists of five base MSA layers followed by a feature pyramid network ~\cite{lin2017feature} module with five MSA layers (scale factor~$=2$). It processes complete video-wise feature embeddings from the frozen pretrained visual encoder. The class categorization is based on sample counts: head classes ($>\!10,\!000$ samples) and tail classes ($<\!1,\!000$ samples). We employ a pre-trained CLIP \cite{radford2021learning} as the foundation model to extract text and visual embeddings for knowledge construction and matching. All models are trained for 800 epochs using SGD with momentum 0.95 and initial learning rate $5\!\times\!10^{-2}$, implemented on a single NVIDIA RTX 3090 GPU for both training and inference. {The average inference time is 2.61s per video, measured on a sequence containing 1,867 frames with an input resolution of $224 \times 224$.} We select the hyperparameters $\lambda = 0.1$ and $\alpha = 0.1$ as they emerge as optimal choices based on the sensitivity analysis (Sec. \ref{sec:sens}).
\begin{table}[t]
\centering
\caption{{Quantitative evaluation on CholecT50 dataset.}}
\label{tab:cholect50_comparison}
\setlength{\tabcolsep}{0.9mm}
\resizebox{\linewidth}{!}{
\begin{tabular}{lcccccc}
\toprule
\textbf{Method} & $\boldsymbol{AP_I}$ & $\boldsymbol{AP_V}$ & $\boldsymbol{AP_T}$ & $\boldsymbol{AP_{IV}}$ & $\boldsymbol{AP_{IT}}$ & $\boldsymbol{AP_{IVT}}$ \\
\midrule
{TCN\cite{nwoye2020recognition}} 
& 48.9 & 29.4 & 21.4 & 17.7& 15.5& 12.4\\
{MLT\cite{nwoye2020recognition}} 
& 84.5 & 48.4 & 28.2& 26.6 & 21.2 & 17.6\\
{Tripnet \cite{nwoye2020recognition}} 
& 92.1 & 54.5& 33.2& 29.7& 26.4& 20.0\\
{RDV \cite{nwoye2022rendezvous}}
& 92.0& 60.7& 38.3& 39.4& 36.9& 29.9\\
{MPCNet \cite{li2024surgical}} 
& 90.9&  62.1&  41.0&  39.0&  36.1& 30.2 \\
{Forest GCN \cite{xi2022forest}} 
& 93.1& 60.1& 40.2& 36.2& 37.5& 36.7\\
{CoT \cite{xi2023chain}}
& \underline{94.1}& 62.5& 41.9& 41.7& 39.5& 38.2\\
\midrule
\textbf{MoeCo-T (Ours)} &{93.7}& \underline{66.2}& \underline{50.4}& \underline{44.1}& \underline{44.1}& \underline{39.5}\\
\textbf{MoeCo-B (Ours)} &\textbf{95.1}& \textbf{69.5}& \textbf{49.7}& \textbf{48.7}& \textbf{46.2}& \textbf{40.5} \\
\bottomrule
\end{tabular}
}
\end{table}
\subsection{Experimental Results}
\subsubsection{Experimental Analysis on Cholect45 Dataset} We compare the proposed framework with eight state-of-the-art (SOTA) triplet recognition methods. As shown in Table \ref{tab:benchmark}, the experimental results highlight the superior performance of our method compared to the competing approaches. In our comparison, MoeCo-T and MoeCo-B denote the utilization of Swin Transformer Tiny (SwinT) and Base (SwinB) as the visual encoder, respectively. Specifically, our MoeCo-T achieves an average $AP_{IVT}$ of 40.5\%, surpassing CurConMix-T by 2.8\% and TERL-T by 4.8\%. Our MoeCo-B achieves a higher performance at 41.7\%, outperforming the second-best method (CurConMix-B) by 2.6\%. For the final prediction, we ensemble the MoeCo-T and MoeCo-B models by averaging their sigmoid probabilities. The resulting model, MoeCo-Ens, achieves the highest performance of 42.6\%, outperforming SelfD by 4.1\% despite SelfD's use of larger backbone networks in its ensemble.

Moreover, our method excels in instrument recognition ($AP_{I}$), instrument-verb association recognition ($AP_{IV}$), and instrument-target association recognition ($AP_{IV}$), demonstrating a clear advantage in tasks related to instruments and supporting the effectiveness of our instrument-anchored knowledge integration. 
Across individual components, our method demonstrates competitive performance, achieving 94.7\%, 72.5\%, and 52.3\% in terms of $AP_{I}$, $AP_{V}$, $AP_{T}$, showing the advantage of our framework in managing multiple tasks. 

\begin{figure}[t]
\centering
\includegraphics[width=8cm]{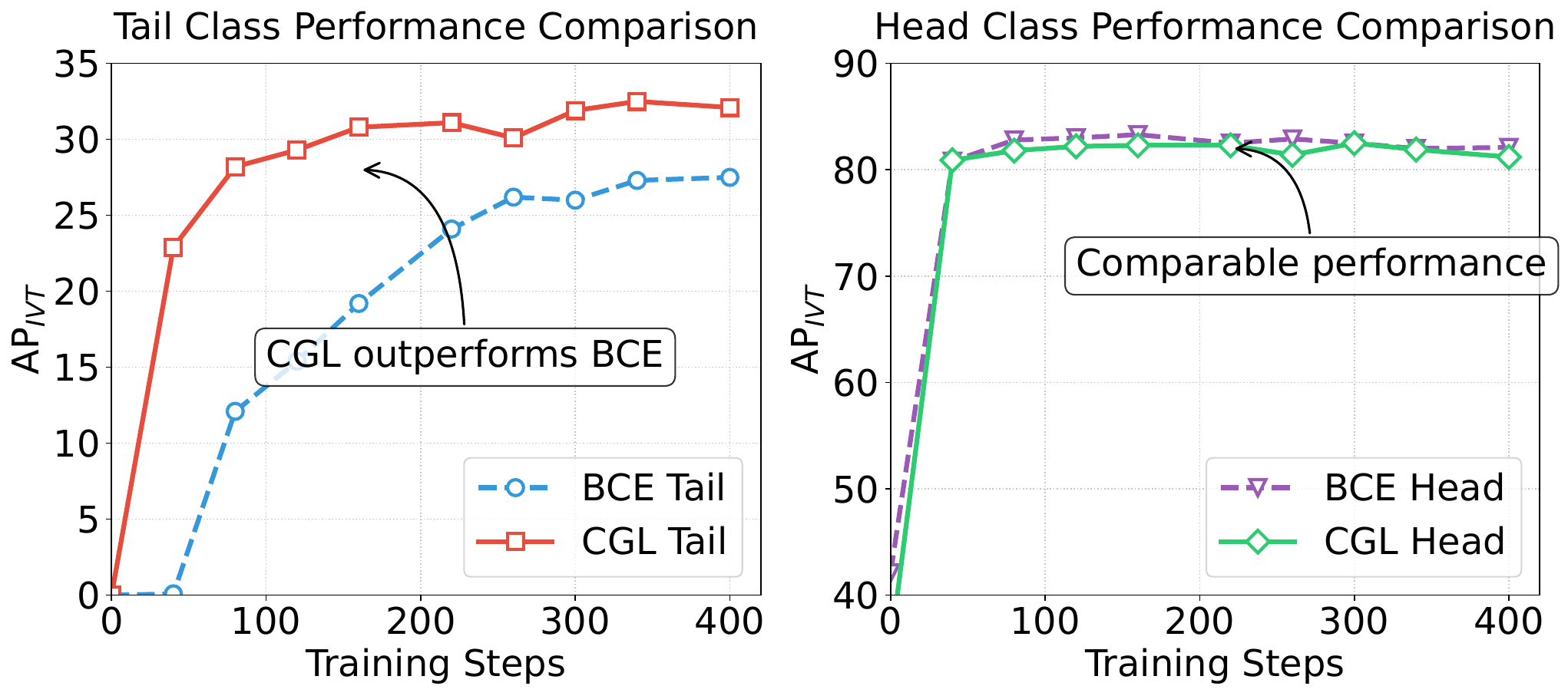}
\caption{Performance comparison between BCE and CGL.}
\label{img:fgl}
\end{figure}
\subsubsection{Experimental Analysis on Cholect50 Dataset} {Table \ref{tab:cholect50_comparison} presents the performance of our method against seven existing approaches under the same RDV data split protocol.} Our proposed methods, MoeCo-B and MoeCo-T, achieve 40.5\% and 39.5\% in $AP_{IVT}$, respectively, with MoeCo-B surpassing the second-best method (CoT) by 2.3\%. Additionally, MoeCo-B consistently demonstrates superior performance across all six metrics compared to other SOTA methods, attaining 95.1\% in $AP_{I}$, 69.5\% in $AP_{V}$, and 49.7\% in $AP_{T}$, highlighting its effectiveness in recognizing individual triplet components as well as their compositional associations.

\subsubsection{Category-Level Performance Analysis} To provide a detailed analysis of per-category performance across the four task metrics, Fig. \ref{fig:box} presents box plots comparing the performance distributions of the baseline method and the proposed MoeCo-T. Overall, MoeCo-T achieves a superior median performance compared to the baseline method across all tasks. More importantly, it demonstrates a substantial elevation in the lower bounds of the distributions, indicating a consistent improvement in recognizing rare (difficult) categories. Furthermore, Fig. \ref{fig:box}(d) reveals a noticeable performance gap between easy and difficult categories in task IVT, underscoring the importance of our framework in enhancing lower-bound performance and contributing to a more balanced and robust overall performance.

{Furthermore, we provide the performance curves of head and tail classes on the validation set during training, as shown in Fig. \ref{img:fgl}.} Under the standard BCE loss, head classes learn substantially faster than tail classes, as the latter are dominated by negative gradients. This imbalance risks overfitting to head classes while leaving tail classes underfitted, leading to suboptimal generalization. In contrast, when CGL is applied to balance positive and negative gradients for tail categories, tail classes converge at a rate comparable to that of head classes. This results in more consistent and coordinated optimization across categories.
\subsection{Ablation Study}

\subsubsection{Ablation on Module Contribution}
As shown in Table \ref{tab:ablation}, the ablation study on the proposed framework clearly demonstrates the individual and synergistic advantages of its three core modules. The baseline model, without any of the proposed modules, achieves 38.3\% in $AP_{IVT}$ score. Introducing the knowledge-driven mixture-of-experts (KD-MoE) mechanism first brings a substantial gain, boosting $AP_{IVT}$ to 40.3\% and improving all individual components by integrating domain knowledge. Subsequently, adding the component-tailored adapter (CTA) further refines the model's ability to capture task-specific features, leading to notable improvements across six metrics. Finally, the incorporation of the coordinated gradient learning (CGL) strategy enhances the perception of rare categories, enabling the full model to achieve the highest scores, culminating in a peak $AP_{IVT}$ of 42.3\%. This progressive enhancement confirms that each module contributes uniquely: KD-MoE enhances relational reasoning and strengthens feature representation, CTA mitigates component-level optimization conflicts, and CGL ensures balanced gradient learning at the category level, with their combination being essential for triplet action recognition.
\begin{table}[t]
\caption{Module contribution ablation of the proposed framework (MoeCo-T). The first row represents the results of the baseline method.}

    \centering
    \fontsize{8pt}{8pt}\selectfont
    % \addtocounter{table}{0}
    %\renewcommand{\arraystretch}{1.2}
    \setlength{\tabcolsep}{0.4mm}{\resizebox{\linewidth}{!}{
\begin{tabular}{@{}ccc|cccccc@{}}
\toprule
\textbf{KD-MoE} & \textbf{CTA} & \textbf{CGL} & $\boldsymbol{AP_I}$ & $\boldsymbol{AP_V}$ & $\boldsymbol{AP_T}$ & $\boldsymbol{AP_{IV}}$ & $\boldsymbol{AP_{IT}}$ & $\boldsymbol{AP_{IVT}}$ \\
\midrule
 {} & {}  & & 89.4 & 67.2 & 53.2 & 42.1 & 45.1 & 38.3 \\
 \checkmark & {}&  & 90.4 & 69.9& 55.4 & 43.3& 47.2 & 40.3 ($\uparrow$2.0)\\
 & \checkmark &  & \underline{91.2} & 67.3 & \underline{57.4} & 41.6 & 48.0 & 40.0 ($\uparrow$1.7)\\
  &  & \checkmark & 91.1 & 70.8 & 55.9 & 43.3& 47.5 & 40.4 ($\uparrow$2.1) \\
 \checkmark & \checkmark  &  & 90.7 & \underline{71.0} & 57.1 & \underline{43.5} & \underline{50.5} & \underline{41.2} ($\uparrow$2.9)\\
 \checkmark & \checkmark  & \checkmark &  \textbf{91.8} & \textbf{72.5} & \textbf{58.0} & \textbf{44.7} & \textbf{50.7} & \textbf{42.3} ($\uparrow$4.0) \\
\bottomrule
\end{tabular}}
}
\label{tab:ablation}
\end{table}
\begin{table}[t]
\centering
\caption{{Ablation study isolating the semantic value of the knowledge base and gating mechanism.}}
\label{tab:random_text}
\resizebox{\linewidth}{!}{\setlength{\tabcolsep}{1.2mm}{
\begin{tabular}{lcccc}
\toprule
\textbf{Text Input Configuration} & $\boldsymbol{AP_I}$ & $\boldsymbol{AP_V}$ & $\boldsymbol{AP_T}$ & $\boldsymbol{AP_{IVT}}$ \\
\midrule
Baseline (Vision-Only)& 89.4 & 67.2 & 53.2 & 38.3 \\
Random Feature Initialization & 89.1 & 67.5 & 54.9 & 38.7 \\
Random Prior Activation & \underline{89.6} & \underline{68.4} & \textbf{55.9} & \underline{39.6} \\
\textbf{w/ KD-MoE (Ours)} & \textbf{90.4} & \textbf{69.9} & \underline{55.4} & \textbf{40.3} \\
\rowcolor{green!10} Oracle Priors (Upper Bound) & 90.7 & 69.1 & 57.3 & 40.9 \\
\bottomrule
\end{tabular}}}
\end{table}

\subsubsection{Ablation on the Semantic Content of KD-MoE} 
{To further isolate the contribution of semantic priors and the gating mechanism, we conducted an additional ablation study under the same backbone, training protocol, and data split. We compare the following variants:
\textbf{(1) Random Feature Initialization}: the MLLM-generated attribute features are replaced with randomly initialized features, removing semantic content while keeping the additional input form;
\textbf{(2) Random Prior Activation}: the original attribute prototypes are retained, but the activated priors are randomly selected instead of being chosen by the learned gating mechanism;
\textbf{(3) Oracle Priors}: ground-truth instrument attributes are used to activate the corresponding priors. This setting is not deployable and is used only as an upper-bound analysis.}

{As shown in Table~\ref{tab:random_text}, replacing the semantic attributes with random features only brings limited improvement over the vision-only baseline, especially for triplet recognition. This indicates that simply adding extra feature inputs is insufficient to explain the gain of KD-MoE. Random prior activation performs better than random feature initialization, suggesting that the MLLM-generated attribute prototypes themselves contain useful domain knowledge and inherently provide a useful regularization effect. However, because the priors are not selected according to the visual content, this variant remains inferior to the full KD-MoE on the primary triplet metric. In contrast, KD-MoE achieves the best deployable performance, improving \(AP_{IVT}\) from 38.3\% to 40.3\%. The oracle setting further improves \(AP_{IVT}\) to 40.9\%, which confirms that correctly activated semantic attributes can provide additional benefit. Importantly, KD-MoE approaches this oracle upper bound without using ground-truth attributes during inference, demonstrating that the learned gating network can effectively select relevant structural priors from visual evidence.}
\begin{figure}[t]
\centering
\includegraphics[width=0.45\textwidth]{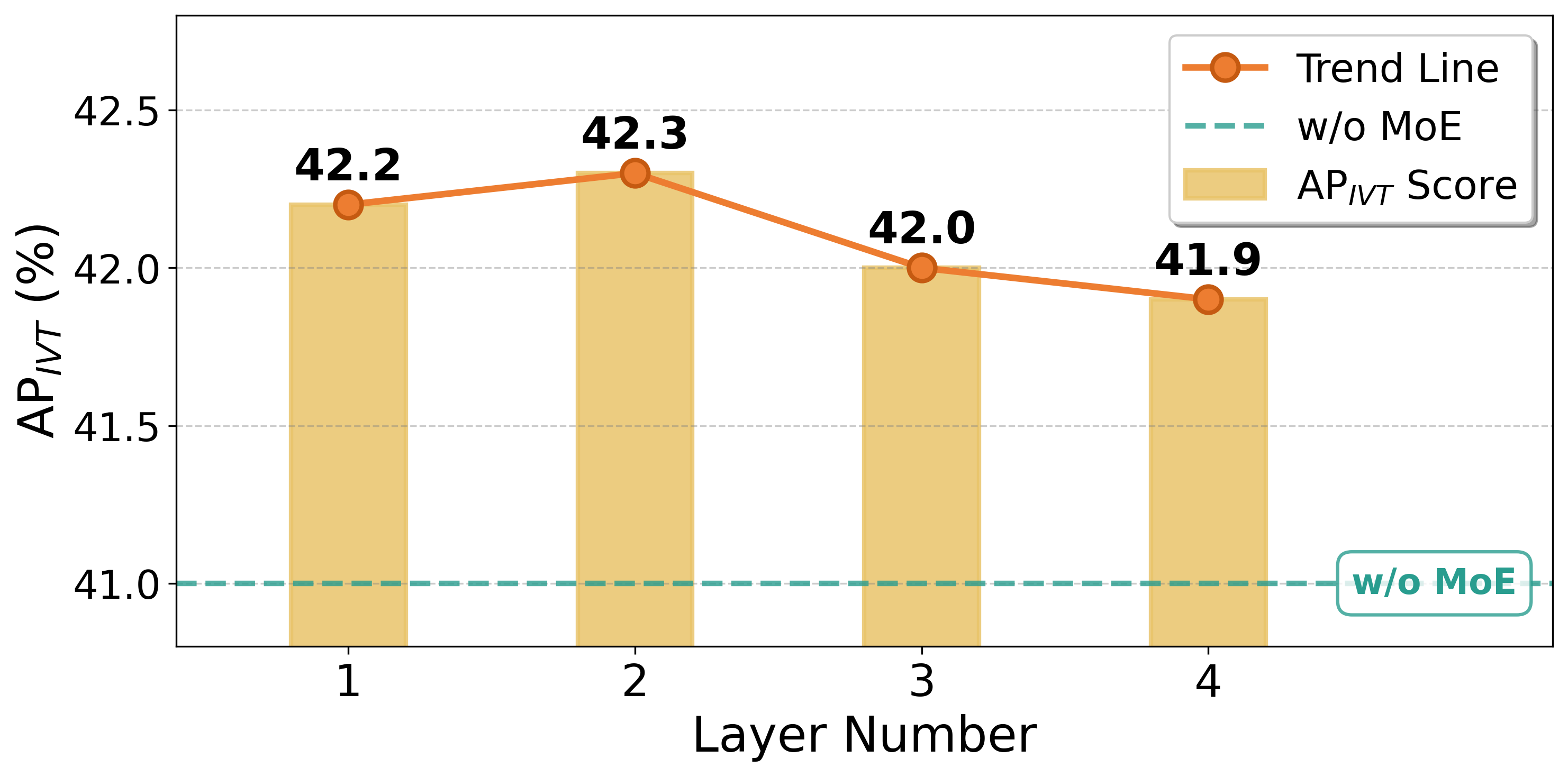}
\caption{KD-MoE ablation at different MSA layers.}
\label{img:moe}
\end{figure}

\begin{figure}[t]
\centering
\includegraphics[width=0.47\textwidth]{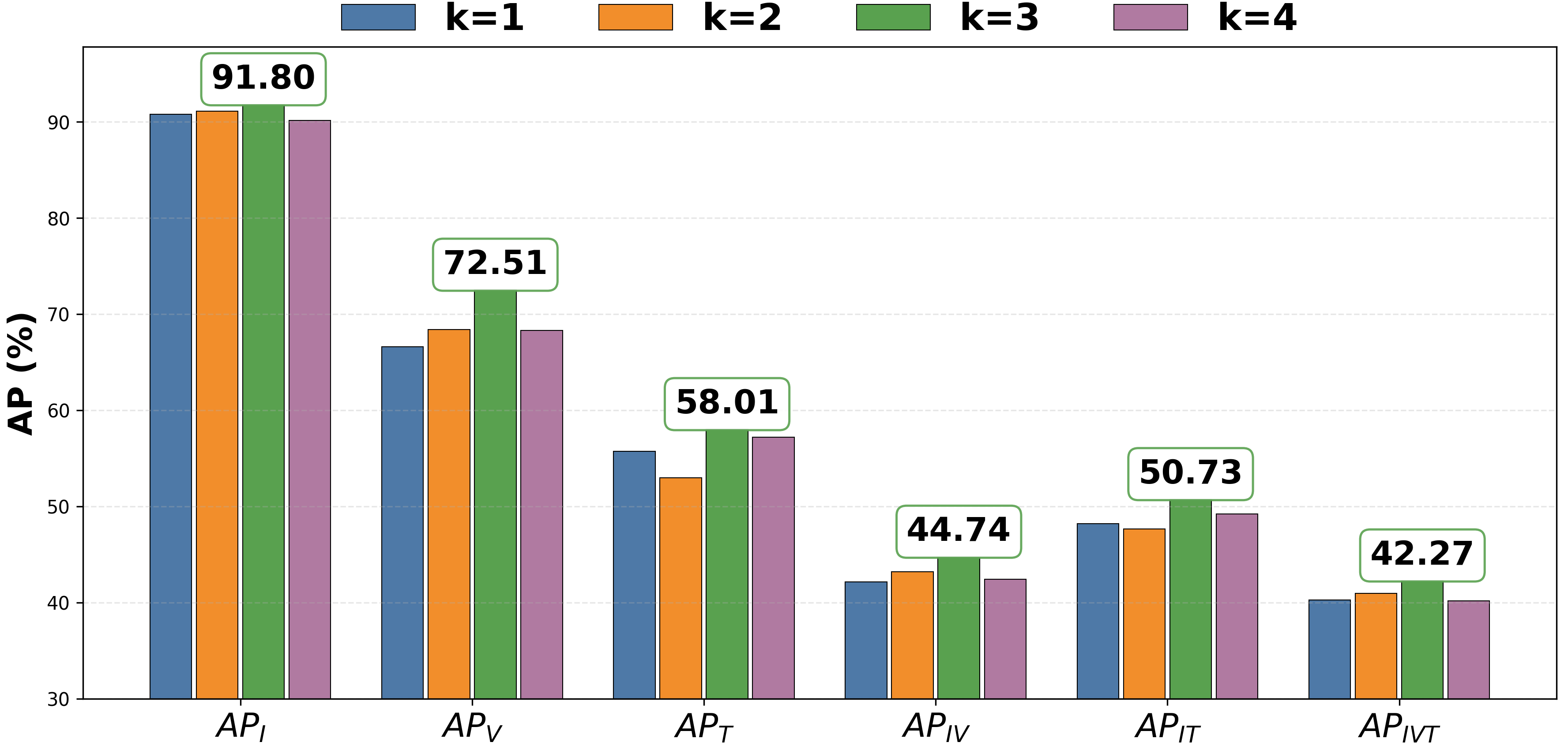}
\caption{Performance under different $k$ options in KD-MoE.}
\label{img:topk}
\end{figure}

\subsubsection{Ablation on KD-MoE Integration Layer}  
Fig. \ref{img:moe} systematically assesses the impact of varying MSA layers for KD-MoE incorporation on the $AP_{IVT}$ metric. The layer number $l \in \{1,2,3,4\}$ denotes the index of the MSA layer integrated within KD-MoE. The results underscore that at $l=2$, the framework achieves the best performance, yielding an $AP_{IVT}$ score of 42.3\%. Remarkably, the $AP_{IVT}$ metric consistently surpasses the performance level observed without KD-MoE integration across diverse layers, showcasing the robustness of the proposed knowledge-driven MoE mechanism when integrated within various feature latent spaces.

\begin{table}[h]
\centering
\caption{{Comparison with representative long-tailed loss functions on CholecT45.}}
\label{tab:loss_comparison}\setlength{\tabcolsep}{1.2mm}{
\resizebox{\linewidth}{!}{
\begin{tabular}{lcccccc}
\toprule
\textbf{Loss Function}
& $\boldsymbol{AP_I}$
& $\boldsymbol{AP_V}$
& $\boldsymbol{AP_T}$
& $\boldsymbol{AP_{IV}}$
& $\boldsymbol{AP_{IT}}$
& $\boldsymbol{AP_{IVT}}$ \\
\midrule
BCE
& 89.4 & 67.2 & 53.2 & 42.1 & 45.1 & 38.3 \\
Focal Loss \cite{lin2017focal}
& 90.2 & \underline{70.5} & \textbf{56.9} & \underline{43.0} & \underline{46.9} & \underline{39.1} \\
CB Loss \cite{cui2019class}
& 90.1 & 68.6 & 55.2 & 42.5 & 46.0 & 38.5 \\
EQ Loss \cite{tan2020equalization}
& \underline{90.9} & 70.2 & 55.6 & 42.6 & 46.3 & 39.0 \\

\textbf{CGL (Ours)}
& \textbf{91.1} & \textbf{70.8} & \underline{55.9} & \textbf{43.3} & \textbf{47.5} & \textbf{40.4} \\
\bottomrule
\end{tabular}
}}
\end{table}
\subsubsection{Ablation on Top-K Activation}
In Eq.~\eqref{equ:topk}, knowledge priors with the top-$k$ probability scores in each expert are activated and integrated with original embeddings. The impact of varying options for $k$ in KD-MoE is illustrated in Fig. \ref{img:topk}. It is evident that there is a notable advantage in each performance metric when $k$ is set to 3. This observation suggests that having 3 attribute descriptions for each instrument structure proves effective for semantic comprehension in surgical scenes.

\subsubsection{Effectiveness of CGL} {To further evaluate the effectiveness of CGL, we compare it with representative long-tailed learning losses, under the same training and evaluation protocol on the SwinT backbone. As shown in Table~\ref{tab:loss_comparison}, CGL achieves the best performance on five out of six evaluation metrics and obtains the highest triplet recognition performance with 40.4\% $AP_{IVT}$. Compared with BCE, Focal Loss, CB Loss, and EQ Loss, CGL improves $AP_{IVT}$ by 2.1\%, 1.3\%, 1.9\%, and 1.4\%, respectively. These results indicate that CGL is more effective in coordinating the optimization dynamics between head and tail categories for surgical triplet recognition.}

\begin{table}[t]
\centering
\caption{{Comparison of computational efficiency and recognition performance.}}
\label{tab:efficiency_comparison}
\resizebox{\linewidth}{!}{\setlength{\tabcolsep}{1mm}{
\begin{tabular}{lcccc}
\toprule
\textbf{Method}
& \textbf{Params}
& \textbf{FLOPs}
& \textbf{Training Time}
& $\boldsymbol{AP_{IVT}}$ \\
\midrule
Baseline
& 95.8M & 4.57G & 19.37s & 38.3 \\
+ 4 Task Branches
  & 108.4M & 4.68G & 19.62s & 38.8 \\
+ VPT \cite{jia2022visual}
  & 95.8M & 4.62G & 19.51s &  39.4\\
+ ST-Adapter \cite{pan2022st}
  & 101.0M & 4.62G & 19.53s  & 39.5\\
+ CTA
  & 100.9M & 4.63G & 19.47s & 40.0 \\
+ KD-MoE
  & 110.8M & 4.60G & 19.40s & 40.3 \\
Ours
  & 116.0M & 4.66G & 19.52s & 41.2 \\
  \bottomrule
  \end{tabular}
  }}
\end{table}
  
\begin{figure*}[h]
\centering
\includegraphics[width=\textwidth]{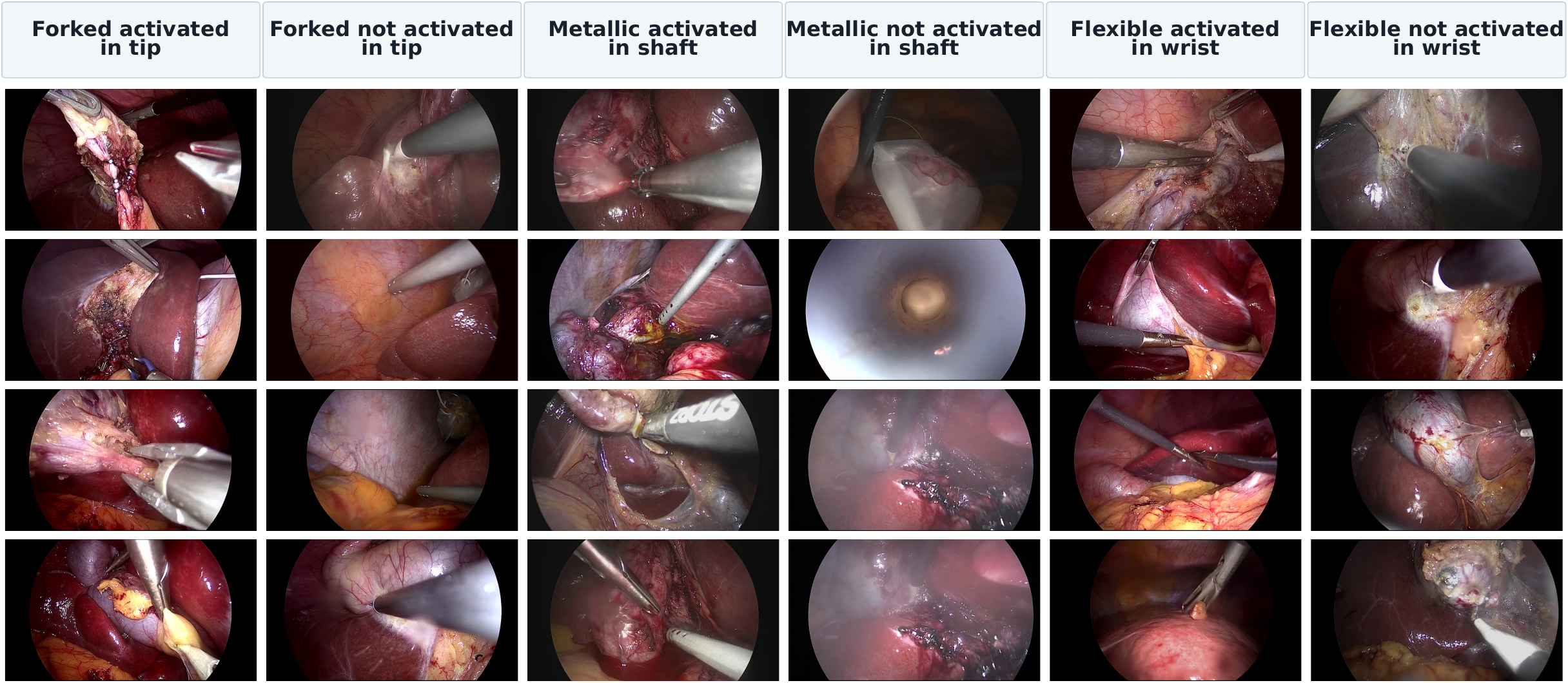}
\caption{{Representative qualitative examples of attribute activation.}}
\label{img:grid}
\end{figure*}

\begin{figure*}[t]
\includegraphics[width=\textwidth]{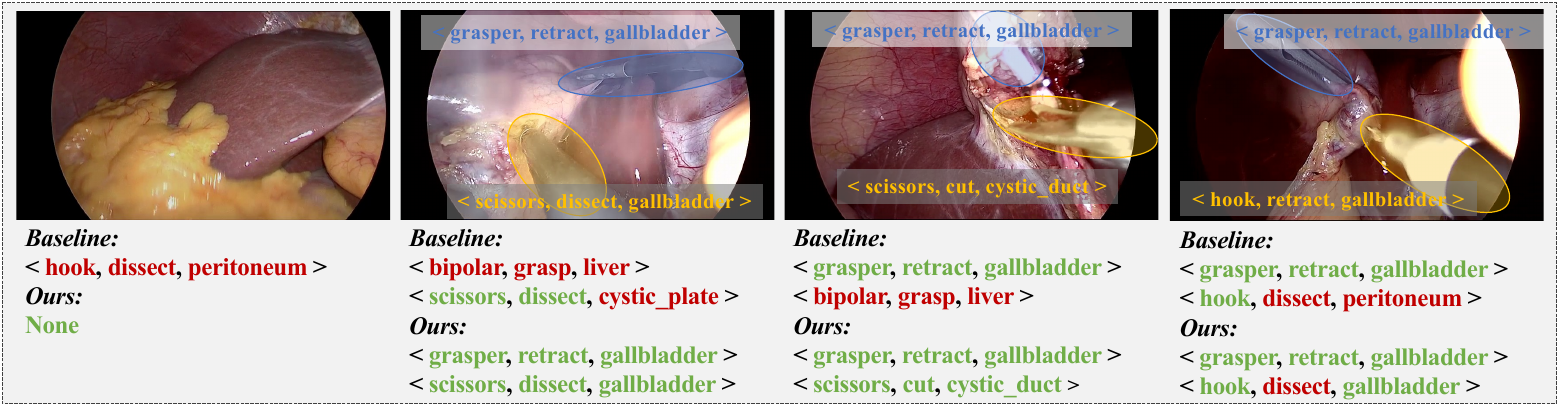}
\caption{Case studies involve the baseline method and our proposed MoeCo-T. Ground truth annotations are displayed on the images. Correct predictions are highlighted in green, and incorrect ones are in red.}
\label{img:vis}
\end{figure*}
\subsection{Computational Efficiency Analysis}
{To quantify the computational overhead of the proposed framework, we report the parameter count, FLOPs per frame, average training time per video, and $AP_{IVT}$ under the same backbone and input setting, as shown in Table~\ref{tab:efficiency_comparison}. Compared with the baseline, the traditional ``4 Task Branches'' design increases the parameter count from 95.8M to 108.4M and the FLOPs from 4.57G to 4.68G, but only improves $AP_{IVT}$ from 38.3\% to 38.8\%. This suggests that simply introducing separate task-specific branches brings additional computational cost but provides limited benefit for triplet recognition. In contrast, CTA achieves a better trade-off between efficiency and accuracy. It increases the parameter count to 100.9M and the FLOPs to 4.63G, while improving $AP_{IVT}$ to 40.0\%. The additional overhead mainly comes from the MSA and MHCA operations used to construct component-tailored representations. Although these attention-based interactions are more computationally involved than simple classifier heads, they enable the model to learn task-specific features for instrument, verb, target, and triplet prediction from shared representations. Compared with VPT and ST-Adapter, CTA has a comparable parameter scale and computational footprint, but achieves higher $AP_{IVT}$, demonstrating that the performance gain benefits from the spatial-temporal separated prompting design rather than merely from increasing learnable parameters.}

{KD-MoE introduces a different type of overhead. Although it evaluates Gaussian likelihoods over attribute components before Top-$K$ expert activation, this computation is performed on compact feature embeddings rather than high-resolution spatial maps. Moreover, the MLLM-derived knowledge base and Gaussian statistics are constructed offline, and the MLLM is not used during training or inference. As a result, KD-MoE increases the FLOPs only slightly from 4.57G to 4.60G and the training time from 19.37s to 19.40s per video, while improving $AP_{IVT}$ from 38.3\% to 40.3\%. When CTA and KD-MoE are combined, the full model achieves the best recognition performance, increasing $AP_{IVT}$ from 38.3\% to 41.2\%, corresponding to an absolute gain of 2.9 percentage points over the baseline. Meanwhile, ours requires 4.66G FLOPs per frame and 19.52s training time per video, which are only 0.09G and 0.15s higher than the baseline, respectively. These results show that the proposed framework achieves a favorable balance between recognition accuracy and computational efficiency.}

\subsection{Visualization}
\label{vis}
{To further substantiate our work visually, we extracted the specific attributes activated for individual predictions. Fig.~\ref{img:grid} presents representative qualitative examples. Each column shows four frames corresponding to one attribute condition, including activated and non-activated cases for tip, shaft, and wrist attributes. The model successfully activates the ``forked'' attribute for instrument tips, ``metallic'' for shafts, and ``flexible'' for wrists only when these visual structures are actually present. In visually inconsistent cases, these attributes are suppressed. This provides strong qualitative evidence that the gating mechanism responds to meaningful visual cues rather than selecting attributes randomly.}
\begin{figure}[t]
\centering
\includegraphics[width=7cm]{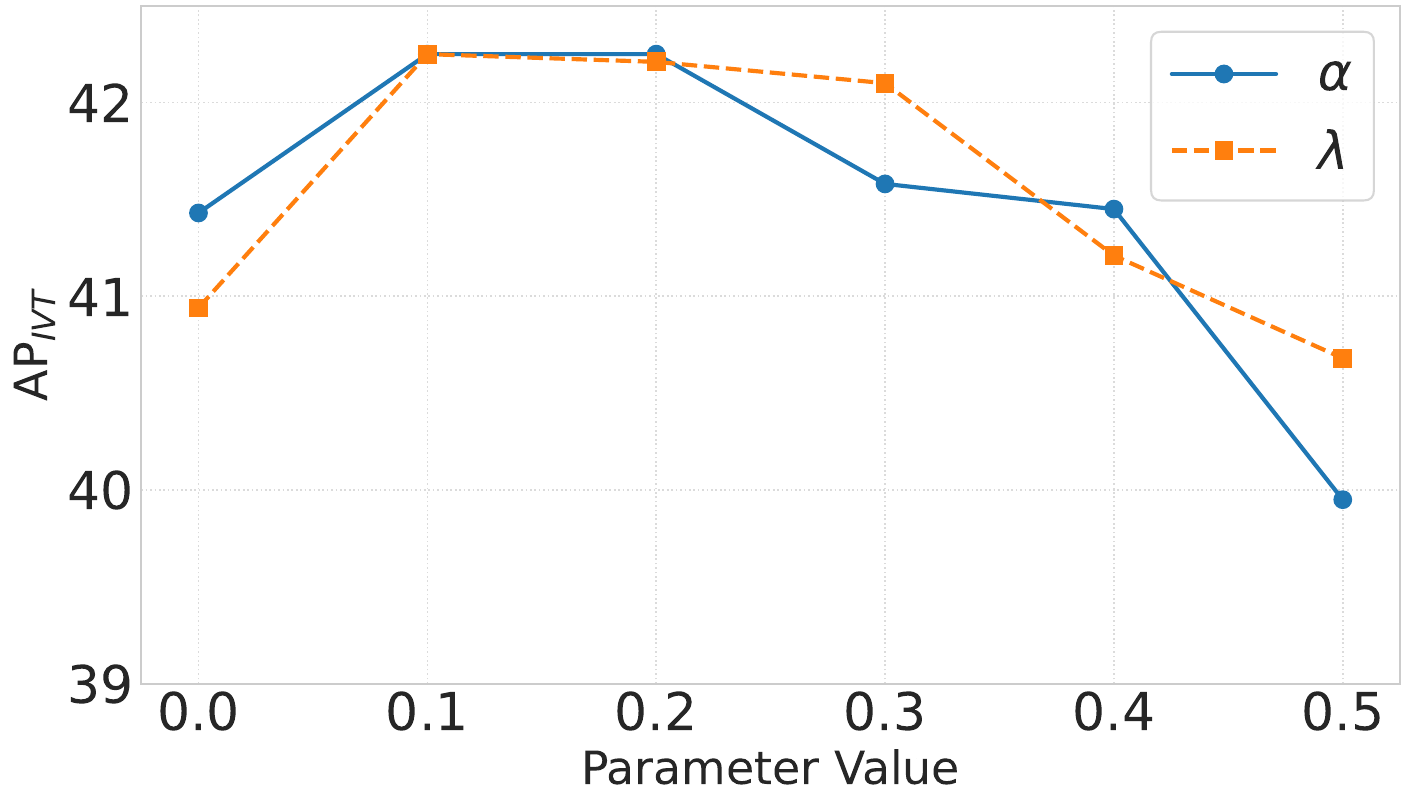}
\caption{$AP_{IVT}$ under different fusion weight of $\alpha$ and $\lambda$.}
\label{img:sens}
\end{figure}

In addition, Fig. \ref{img:vis} showcases several samples of complex surgical scenarios, highlighting challenges such as smoke and motion blur. Among the three triplet components, instruments play a pivotal role as they serve as the initial step in understanding ongoing procedures and exhibit the most distinguishable visual features for identification. This underscores the importance of leveraging instruments to guide the recognition of other sub-tasks and eliminate false positive results, as shown in the first case. Compared to the baseline, the proposed method demonstrates significant improvements in recognizing each component. However, limitations remain, as evidenced by failure cases, such as confusion between verbs ``dissect'' and ``retract'', where the minimal visual or semantic differences between them pose a challenge for precise distinction.

\subsection{Sensitivity Analysis}
\label{sec:sens}
\subsubsection{Hyperparameters Selection}
We examine two hyperparameters: 1) A weighted parameter $\alpha$ utilized in CTA, which functions to balance the contribution of task-specific features integrated into the singular triplet task in Eq. \eqref{eq:zivt}. 2) A probabilistic variable $\lambda$ within CGL. Eq.~\eqref{eq:cgl} utilizes $\lambda$ as the probability for dropping affected loss elements. Specifically, we set $\lambda = 0.1$ and explore $\alpha$ in the range $\{0, 0.1, 0.2, 0.3, 0.4, 0.5\}$. Subsequently, we vary $\lambda$ across $\{0, 0.1, 0.2, 0.3, 0.4, 0.5\}$ while maintaining $\alpha = 0.1$. Fig. \ref{img:sens} demonstrates the stability of $AP_{IVT}$ scores between 0.1 and 0.2, with $\lambda = 0.1$ and $\alpha = 0.1$ identified as optimal choices. Results indicate that dropping gradients with a probability of $\lambda = 0.1$ significantly enhances the performance, showcasing the effectiveness of CGL. However, excessive removal of positive gradients from head classes can degrade their performance, highlighting the need for a critical point that improves tail classes while preserving head class performance.

\subsubsection{Head/Tail Split Criterion}
{In our primary experiments, the definition of head and tail classes follows the established protocol in TERL \cite{gui2024tail} and aligns with the empirical class-frequency distribution of the CholecT45 dataset, where a natural gap separates highly dominant categories from rare ones. However, to evaluate the robustness of our proposed CGL method and ensure its generalizability to datasets of varying scales, we transition from absolute counts to relative, percentile-based thresholds. For the CholecT45 dataset, our original absolute split (\(>10,000 / <1,000\)) corresponds exactly to a relative split of \(>8\% / <0.8\%\) of the total dataset size. To systematically analyze the impact of these boundaries, we evaluate CGL across three different relative threshold settings: \(>5\% / <0.5\%\), \(>8\% / <0.8\%\), and \(>10\% / <1.0\%\).}

\begin{table}[t]
\centering
\caption{{Sensitivity analysis (${AP_{IVT}}$) of CGL under different relative head/tail split criteria.}}
\label{tab:sens_threshold}
\begin{tabular}{lcccc}
\toprule
\textbf{Split Criterion} & $\boldsymbol{AP_I}$
& $\boldsymbol{AP_V}$
& $\boldsymbol{AP_T}$
& $\boldsymbol{AP_{IVT}}$ \\
\midrule
Ours w/o CGL &90.7 & {71.0} & 57.1 & 41.2 \\
$>5\%/<0.5\%$ & 91.0 & 71.8 & 57.6 & 41.6  \\
$>8\%/<0.8\%$ & \textbf{91.8} & \textbf{72.5} & \textbf{58.0} & \textbf{42.3}  \\
$>10\%/<1.0\%$ & 91.3 & \textbf{72.5} & 57.7 & 42.0  \\
\bottomrule
\end{tabular}
\end{table}

{As shown in Table \ref{tab:sens_threshold}, the overall performance remains highly stable across the different threshold settings, fluctuating only slightly between 41.6\% and 42.3\%. Importantly, all three configurations outperform the baseline without CGL (41.2\%), demonstrating the robustness of the proposed method.
The best performance is obtained with the $>8\%/<0.8\%$ criterion, which corresponds to the original split used in our experiments. These results indicate that CGL is not highly sensitive to a specific head/tail boundary and can robustly improve optimization under different reasonable split criteria. Furthermore, directly transferring the thresholds from CholecT45 to CholecT50 without tuning still yielded strong results, suggesting the inherent robustness of CGL across different data distributions.} 

{Nevertheless, to ensure generalizability across entirely new datasets of varying scales, we recommend defining head and tail classes based on relative class frequencies. In practice, classes with very high relative frequencies are treated as head classes, while those with extremely low relative frequencies are treated as tail classes. The remaining categories are regarded as medium classes and are not explicitly modulated by CGL. Aligning the head/tail boundaries with these distributional gaps provides an adaptive and dataset-size-independent strategy. Based on our sensitivity analysis, relative thresholds within 5\%--10\% for head classes and 0.5\%--1.0\% for tail classes can serve as reasonable initial settings for new datasets.}

\section{Conclusion}
In conclusion, this paper introduced \textit{MoeCo}, a novel mixture-of-experts-guided co-optimization framework for surgical triplet recognition. By addressing component-level conflicts with a component-tailored adapter and mitigating category-level competition with a coordinated gradient learning strategy, \textit{MoeCo} effectively resolves hierarchical optimization conflicts at both levels. Crucially, the integration of a knowledge-driven mixture-of-experts mechanism, inspired by domain priors, further enhances the model with more expressive representations. However, the limited availability of public annotated data for diverse surgical scenarios is a bottleneck. Future endeavors could focus on cultivating contextual understanding with MLLMs to enhance zero-shot application and facilitate broader usage across various scenarios.


\begin{thebibliography}{10}

\bibitem{jin2021temporal}
Y.~Jin, Y.~Long, C.~Chen, Z.~Zhao, Q.~Dou, and P.-A. Heng, ``Temporal memory relation network for workflow recognition from surgical video,'' {\em IEEE Transactions on Medical Imaging}, vol.~40, no.~7, pp.~1911--1923, 2021.

\bibitem{guo2025surgical}
D.~Guo, W.~Si, Z.~Li, J.~Pei, and P.-A. Heng, ``Surgical workflow recognition and blocking effectiveness detection in laparoscopic liver resection with pringle maneuver,'' in {\em Conference on Artificial Intelligence}, vol.~39, pp.~3220--3228, 2025.

\bibitem{yang2025surgpetl}
S.~Yang, Z.~Cai, L.~Luo, N.~Ma, S.~Xu, and H.~Chen, ``Surgpetl: Parameter-efficient image-to-surgical-video transfer learning for surgical phase recognition,'' {\em IEEE Transactions on Medical Imaging}, 2025.

\bibitem{dagnino2024robot}
G.~Dagnino and D.~Kundrat, ``Robot-assistive minimally invasive surgery: trends and future directions,'' {\em International Journal of Intelligent Robotics and Applications}, vol.~8, no.~4, pp.~812--826, 2024.

\bibitem{zheng2025survey}
Y.~Zheng {\em et~al.}, ``A survey of embodied learning for object-centric robotic manipulation,'' {\em Machine Intelligence Research}, pp.~1--39, 2025.

\bibitem{moglia2021systematic}
A.~Moglia, K.~Georgiou, E.~Georgiou, R.~M. Satava, and A.~Cuschieri, ``A systematic review on artificial intelligence in robot-assisted surgery,'' {\em International Journal of Surgery}, vol.~95, p.~106151, 2021.

\bibitem{nwoye2020recognition}
C.~I. Nwoye {\em et~al.}, ``Recognition of instrument-tissue interactions in endoscopic videos via action triplets,'' in {\em International Conference on Medical Image Computing and Computer-Assisted Intervention}, pp.~364--374, 2020.

\bibitem{nwoye2023cholectriplet2021}
C.~I. Nwoye {\em et~al.}, ``Cholectriplet2021: A benchmark challenge for surgical action triplet recognition,'' {\em Medical Image Analysis}, vol.~86, p.~102803, 2023.

\bibitem{li2023mt}
Y.~Li, T.~Xia, H.~Luo, B.~He, and F.~Jia, ``Mt-fist: a multi-task fine-grained spatial-temporal framework for surgical action triplet recognition,'' {\em IEEE journal of biomedical and health informatics}, vol.~27, no.~10, pp.~4983--4994, 2023.

\bibitem{gui2024tail}
S.~Gui and Z.~Wang, ``Tail-enhanced representation learning for surgical triplet recognition,'' in {\em International Conference on Medical Image Computing and Computer-Assisted Intervention}, pp.~689--699, 2024.

\bibitem{peiinstrument}
J.~Pei, J.~Zhang, G.~Qin, K.~Wang, Y.~Jin, and P.-A. Heng, ``Instrument-tissue-guided surgical action triplet detection via textual-temporal trail exploration,'' {\em IEEE Transactions on Medical Imaging}, 2025.

\bibitem{nwoye2022rendezvous}
C.~I. Nwoye {\em et~al.}, ``Rendezvous: Attention mechanisms for the recognition of surgical action triplets in endoscopic videos,'' {\em Medical Image Analysis}, vol.~78, p.~102433, 2022.

\bibitem{liu2024surgical}
J.~Liu {\em et~al.}, ``Surgical action triplet recognition assisted by foundation models-based instrument localization,'' in {\em International Conference on Neural Information Processing}, pp.~383--396, 2024.

\bibitem{sharma2023rendezvous}
S.~Sharma, C.~I. Nwoye, D.~Mutter, and N.~Padoy, ``Rendezvous in time: an attention-based temporal fusion approach for surgical triplet recognition,'' {\em International Journal of Computer Assisted Radiology and Surgery}, vol.~18, no.~6, pp.~1053--1059, 2023.

\bibitem{jung2021towards}
H.~Jung and Y.~Oh, ``Towards better explanations of class activation mapping,'' in {\em IEEE International Conference on Computer Vision}, pp.~1336--1344, 2021.

\bibitem{ban2024fair}
H.~Ban and K.~Ji, ``Fair resource allocation in multi-task learning,'' in {\em International Conference on Machine Learning}, pp.~2715--2731, 2024.

\bibitem{luo2017vision}
X.~Luo, A.~J. McLeod, S.~E. Pautler, C.~M. Schlachta, and T.~M. Peters, ``Vision-based surgical field defogging,'' {\em IEEE Transactions on Medical Imaging}, vol.~36, no.~10, pp.~2021--2030, 2017.

\bibitem{pei2025restore}
J.~Pei {\em et~al.}, ``Benchmarking laparoscopic surgical image restoration and beyond,'' {\em arXiv preprint arXiv:2505.19161}, 2025.

\bibitem{hurst2024gpt}
A.~Hurst {\em et~al.}, ``Gpt-4o system card,'' {\em arXiv:2410.21276}, 2024.

\bibitem{nwoye2022data}
C.~I. Nwoye and N.~Padoy, ``Data splits and metrics for method benchmarking on surgical action triplet datasets,'' {\em arXiv:2204.05235}, 2022.

\bibitem{li2024surgical}
Y.~Li, Z.~Zhao, and R.~Li, ``Surgical action triplet recognition by using a multi-task prior-reinforced and cross-sample network,'' in {\em International Conference on Electronic Engineering and Informatics}, pp.~1283--1289, 2024.

\bibitem{estabrooks2004multiple}
A.~Estabrooks, T.~Jo, and N.~Japkowicz, ``A multiple resampling method for learning from imbalanced data sets,'' {\em Computational Intelligence}, vol.~20, no.~1, pp.~18--36, 2004.

\bibitem{zheng2020deep}
Y.~Zheng, H.~Yao, and X.~Sun, ``Deep semantic parsing of freehand sketches with homogeneous transformation, soft-weighted loss, and staged learning,'' {\em IEEE Transactions on Multimedia}, vol.~23, pp.~3590--3602, 2020.

\bibitem{lin2024distributionally}
D.~Lin, T.~Peng, R.~Chen, X.~Xie, X.~Qin, and Z.~Cui, ``Distributionally robust loss for long-tailed multi-label image classification,'' in {\em European Conference on Computer Vision}, pp.~417--433, 2024.

\bibitem{gui2024mt4mtl}
S.~Gui, Z.~Wang, J.~Chen, X.~Zhou, C.~Zhang, and Y.~Cao, ``Mt4mtl-kd: A multi-teacher knowledge distillation framework for triplet recognition,'' {\em IEEE Transactions on Medical Imaging}, vol.~43, no.~4, pp.~1628--1639, 2024.

\bibitem{jacobs1991adaptive}
R.~A. Jacobs, M.~I. Jordan, S.~J. Nowlan, and G.~E. Hinton, ``Adaptive mixtures of local experts,'' {\em Neural Computation}, vol.~3, no.~1, pp.~79--87, 1991.

\bibitem{shazeer2017outrageously}
N.~Shazeer {\em et~al.}, ``Outrageously large neural networks: The sparsely-gated mixture-of-experts layer,'' {\em arXiv:1701.06538}, 2017.

\bibitem{guo2018dynamic}
M.~Guo, A.~Haque, D.-A. Huang, S.~Yeung, and L.~Fei-Fei, ``Dynamic task prioritization for multitask learning,'' in {\em European Conference on Computer Vision}, pp.~270--287, 2018.

\bibitem{fedus2022switch}
W.~Fedus, B.~Zoph, and N.~Shazeer, ``Switch transformers: Scaling to trillion parameter models with simple and efficient sparsity,'' {\em Journal of Machine Learning Research}, vol.~23, no.~120, pp.~1--39, 2022.

\bibitem{radford2021learning}
A.~Radford {\em et~al.}, ``Learning transferable visual models from natural language supervision,'' in {\em International Conference on Machine Learning}, pp.~8748--8763, 2021.

\bibitem{najar2017comparison}
F.~Najar, S.~Bourouis, N.~Bouguila, and S.~Belghith, ``A comparison between different gaussian-based mixture models,'' in {\em IEEE International Conference on Computer Systems and Applications}, pp.~704--708, 2017.

\bibitem{dai2024deepseekmoe}
D.~Dai {\em et~al.}, ``Deepseekmoe: Towards ultimate expert specialization in mixture-of-experts language models,'' {\em arXiv:2401.06066}, 2024.

\bibitem{zhou2022learning}
K.~Zhou, J.~Yang, C.~C. Loy, and Z.~Liu, ``Learning to prompt for vision-language models,'' {\em International Journal of Computer Vision}, vol.~130, no.~9, pp.~2337--2348, 2022.

\bibitem{tan2020equalization}
J.~Tan {\em et~al.}, ``Equalization loss for long-tailed object recognition,'' in {\em IEEE Conference on Computer Vision and Pattern Recognition}, pp.~11662--11671, 2020.

\bibitem{chen2023surgical}
Y.~Chen, S.~He, Y.~Jin, and J.~Qin, ``Surgical activity triplet recognition via triplet disentanglement,'' in {\em International Conference on Medical Image Computing and Computer-Assisted Intervention}, pp.~451--461, 2023.

\bibitem{yamlahi2023self}
A.~Yamlahi {\em et~al.}, ``Self-distillation for surgical action recognition,'' in {\em International Conference on Medical Image Computing and Computer-Assisted Intervention}, pp.~637--646, 2023.

\bibitem{jeon2025curconmix}
Y.~Jeon {\em et~al.}, ``Curconmix: A curriculum contrastive learning framework for enhancing surgical action triplet recognition,'' in {\em International Conference on Medical Image Computing and Computer-Assisted Intervention}, pp.~149--158, 2025.

\bibitem{li2024parameter}
Y.~Li, B.~Bai, and F.~Jia, ``Parameter-efficient framework for surgical action triplet recognition,'' {\em International Journal of Computer Assisted Radiology and Surgery}, pp.~1--9, 2024.

\bibitem{liu2021swin}
Z.~Liu {\em et~al.}, ``Swin transformer: Hierarchical vision transformer using shifted windows,'' in {\em IEEE International Conference on Computer Vision}, pp.~10012--10022, 2021.

\bibitem{zhang2022actionformer}
C.-L. Zhang, J.~Wu, and Y.~Li, ``Actionformer: Localizing moments of actions with transformers,'' in {\em European Conference on Computer Vision}, pp.~492--510, 2022.

\bibitem{lin2017feature}
T.-Y. Lin, P.~Doll{\'a}r, R.~Girshick, K.~He, B.~Hariharan, and S.~Belongie, ``Feature pyramid networks for object detection,'' in {\em IEEE Conference on Computer Vision and Pattern Recognition}, pp.~2117--2125, 2017.

\bibitem{xi2022forest}
N.~Xi, J.~Meng, and J.~Yuan, ``Forest graph convolutional network for surgical action triplet recognition in endoscopic videos,'' {\em IEEE Transactions on Circuits and Systems for Video Technology}, vol.~32, no.~12, pp.~8550--8561, 2022.

\bibitem{xi2023chain}
N.~Xi, J.~Meng, and J.~Yuan, ``Chain-of-look prompting for verb-centric surgical triplet recognition in endoscopic videos,'' in {\em ACM International Conference on Multimedia}, pp.~5007--5016, 2023.

\bibitem{lin2017focal}
T.-Y. Lin, P.~Goyal, R.~Girshick, K.~He, and P.~Doll{\'a}r, ``Focal loss for dense object detection,'' in {\em IEEE International Conference on Computer Vision}, pp.~2980--2988, 2017.

\bibitem{cui2019class}
Y.~Cui, M.~Jia, T.-Y. Lin, Y.~Song, and S.~Belongie, ``Class-balanced loss based on effective number of samples,'' in {\em IEEE/CVF Conference on Computer Vision and Pattern Recognition}, pp.~9268--9277, 2019.

\bibitem{jia2022visual}
M.~Jia, L.~Tang, B.-C. Chen, C.~Cardie, S.~Belongie, B.~Hariharan, and S.-N. Lim, ``Visual prompt tuning,'' in {\em European conference on computer vision}, pp.~709--727, Springer, 2022.

\bibitem{pan2022st}
J.~Pan, Z.~Lin, X.~Zhu, J.~Shao, and H.~Li, ``St-adapter: Parameter-efficient image-to-video transfer learning,'' {\em Advances in Neural Information Processing Systems}, vol.~35, pp.~26462--26477, 2022.

\end{thebibliography}
\end{document}